\documentclass[10pt,twocolumn,letterpaper]{article}

\usepackage[pagenumbers]{cvpr}

\newcommand{\TODO}[1]{\textbf{\color{red}[TODO: #1]}}

\usepackage{float}

\definecolor{cvprblue}{rgb}{0.21,0.49,0.74}
\usepackage[pagebackref,breaklinks,colorlinks,allcolors=cvprblue]{hyperref}

\usepackage{graphicx}
\usepackage{booktabs}
\usepackage{soul}
\usepackage{balance}

\usepackage{comment}

\iftrue
	\definecolor{jtcolor}{RGB}{0,0,255}
	\newcommand\JT[1] {\emph{\textcolor{jtcolor}{JT: #1}}}

	\definecolor{todocolor}{RGB}{255,0,00}

	\definecolor{wzcolor}{RGB}{163, 159, 225}
	\newcommand\WZ[1] {\emph{\textcolor{wzcolor}{WZ: #1}}}
\else 
	\newcommand\JT[1] {}
	\newcommand\TODO[1] {}
	\newcommand\WZ[2] {}
\fi 

\usepackage{pifont} 
\usepackage{booktabs}
\usepackage{multirow}
\usepackage{adjustbox}
\usepackage{fontawesome} 
\usepackage{amsmath} 

\newcommand{\norm}[1]{\left\lVert#1\right\rVert}
\definecolor{opA}{rgb}{0.9,0.6,0.0}
\definecolor{opB}{rgb}{0.35,0.70,0.90}
\definecolor{opC}{rgb}{0.8,0.40,0.0}
\definecolor{opD}{rgb}{0.0,0.60,0.50} %
\definecolor{opE}{rgb}{0.8,0.6,0.7}
\definecolor{opF}{rgb}{0.,0.45,0.70} 

\definecolor{pltBlue}{rgb}{0.12156862745098039, 0.4666666666666667, 0.7058823529411765}
\definecolor{pltOrange}{rgb}{1.0, 0.4980392156862745, 0.054901960784313725}
\definecolor{pltGreen}{rgb}{0.17254901960784313, 0.6274509803921569, 0.17254901960784313}
\definecolor{pltRed}{rgb}{0.8392156862745098, 0.15294117647058825, 0.1568627450980392}
\definecolor{pltViolet}{rgb}{0.5803921568627451, 0.403921568627451, 0.7411764705882353}
\definecolor{pltBrown}{rgb}{0.5490196078431373, 0.33725490196078434, 0.29411764705882354}
\definecolor{pltMagenta}{rgb}{0.8901960784313725, 0.4666666666666667, 0.7607843137254902}
\definecolor{pltGray}{rgb}{0.4980392156862745, 0.4980392156862745, 0.4980392156862745}
\definecolor{pltLightGreen}{rgb}{0.7372549019607844, 0.7411764705882353, 0.13333333333333333}
\definecolor{pltCyan}{rgb}{0.09019607843137255, 0.7450980392156863, 0.8117647058823529}
\definecolor{pltPink}{rgb}{0.49803921568627, 0.49803921568627, 0.49803921568627}

\definecolor{cmarkcolor}{rgb}{0.49,0.74,0.49}
\definecolor{xmarkcolor}{rgb}{0.86,0.34,0.34}

\newcommand{\xmark}{\textcolor{xmarkcolor}{\ding{55}}}

\def\link#1{
    \ifx&#1&
        \xmark{}
    \else
        {\href{#1}{\faExternalLink}}
    \fi
}

\newcommand{\shortname}{GEM\xspace}
\newcommand{\longname}{Gaussian Eigen Models for Human Heads\xspace}

\renewcommand{\paragraph}[1]{\noindent\textbf{#1}}

\title{Gaussian Eigen Models for Human Heads}

\author{
Wojciech Zielonka\textsuperscript{1, 2} \quad Timo Bolkart\textsuperscript{3} \quad Thabo Beeler\textsuperscript{3} \quad Justus Thies\textsuperscript{1, 2} \\ \\
\textsuperscript{1}Max Planck Institute for Intelligent Systems, Tübingen, Germany \\
\textsuperscript{2}Technical University of Darmstadt \quad
\textsuperscript{3}Google \\
\url{https://zielon.github.io/gem/}
}

\begin{document}


\twocolumn[{%
\renewcommand\twocolumn[1][]{#1}%
\maketitle
\begin{center}
    \centering
    \captionsetup{type=figure}
    \includegraphics[width=1.0\textwidth]{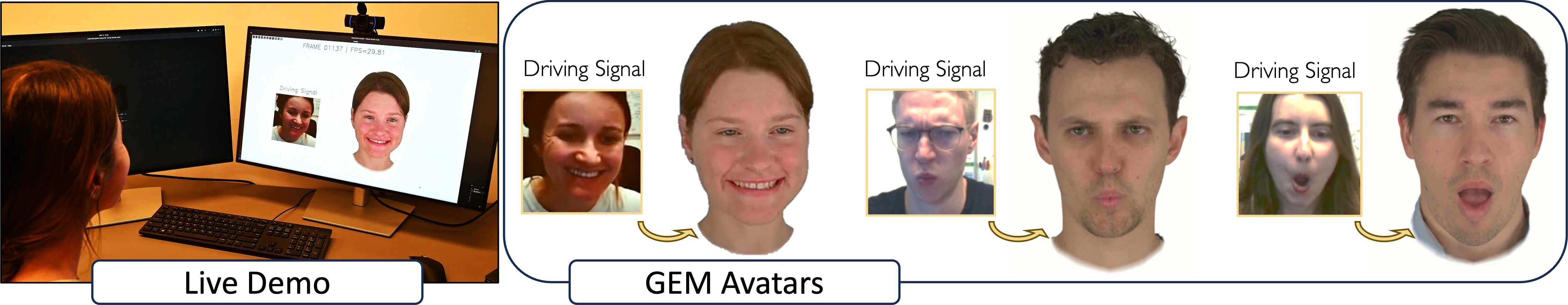}
    \caption{We propose a method that represents 3D Gaussian head avatars in a network-free form as ensembles of eigenbases (GEM). Only a linear combination of these bases is needed to generate new primitives, which can be splatted using 3D Gaussian Splatting. We demonstrate that the necessary coefficients for a specific expression can be regressed from single images, enabling real-time facial animation and cross-reenactment. The simplicity of GEM results in highly efficient storage and rendering times.}
    \label{fig:teaser}
\end{center}%
}]

\begin{abstract}
Current personalized neural head avatars face a trade-off: lightweight models lack detail and realism, while high-quality, animatable avatars require significant computational resources, making them unsuitable for commodity devices.
To address this gap, we introduce Gaussian Eigen Models (\shortname), which provide high-quality, lightweight, and easily controllable head avatars.
\shortname utilizes 3D Gaussian primitives for representing the appearance combined with Gaussian splatting for rendering.
Building on the success of mesh-based 3D morphable face models (3DMM), we define \shortname as an ensemble of linear eigenbases for representing the head appearance of a specific subject.
In particular, we construct linear bases to represent the position, scale, rotation, and opacity of the 3D Gaussians.
This allows us to efficiently generate Gaussian primitives of a specific head shape by a linear combination of the basis vectors, only requiring a low-dimensional parameter vector that contains the respective coefficients.
We propose to construct these linear bases (\shortname) by distilling high-quality compute-intense CNN-based Gaussian avatar models that can generate expression-dependent appearance changes like wrinkles.
These high-quality models are trained on multi-view videos of a subject and are distilled using a series of principle component analyses.
Once we have obtained the bases that represent the animatable appearance space of a specific human, we learn a regressor that takes a single RGB image as input and predicts the low-dimensional parameter vector that corresponds to the shown facial expression.
We demonstrate that this regressor can be trained such that it effectively supports self- and cross-person reenactment from monocular videos without requiring prior mesh-based tracking.
In a series of experiments, we compare \shortname's self-reenactment and cross-person reenactment results to state-of-the-art 3D avatar methods, demonstrating \shortname's higher visual quality and better generalization to new expressions.
As our distilled linear model is highly efficient in generating novel animation states, we also show a real-time demo of GEMs driven by monocular webcam videos.
The code and model will be released for research purposes.
\end{abstract}

\section{Introduction}
\label{sec:intro}
Half a century ago, Frederick Parke described a representation and animation technique to generate ,,animated sequences of a human face changing expressions''~\cite{parke1972}.
Using polygonal meshes, single facial expression states were described that could be combined with linear interpolation to generate new expression states (the ,,simplest way consistent with natural motion''~\cite{parke1972}).
Based on this principle, Blanz, and Vetter \cite{Blanz1999AMM} introduced the so-called 3D morphable model (3DMM) - a statistical model of the 3D shape and appearance of human faces.
Principle Component Analysis (PCA) is performed on a set of around 200 subjects that have been laser-scanned and registered to a consistent template to find the displacement vectors (principal components) of how faces change the most, in terms of geometry and albedo.
With this PCA basis, new faces can be generated by specifying the coefficients for the principle components taking a dot product of the coefficients with the basis to obtain offsets, and adding them to the mean.
State-of-the-art reports on face reconstruction and tracking~\cite{zollhoefer2018facestar} as well as on morphable models~\cite{egger2020morphablemodels} state that this representation is widely used for facial performance capturing (regression-based and optimization-based) and builds the backbone of recent controllable photo-realistic 3D avatars that are equipped with neural rendering~\cite{tewari2020neuralrendering,tewari2022advances, Zheng2021IMA, grassal2022neural, Gafni2020DynamicNR}.
Inspired by the simplicity of such mesh-based linear morphable models and addressing the lack of appearance realism of current 3DMMs, we propose a personalized linear appearance model based on 3D Gaussians as geometry primitives following 3D Gaussian Splatting (3DGS)~\cite{Kerbl20233DGS}.
In contrast to the work on Dynamic 3D Gaussian Avatars \cite{Li2023AnimatableGL, Qian2024gaussianavatars, Zielonka2025Drivable3D, Pang2023ASHAG, Zheng2023GPSGaussianGP, Xu2023GaussianHA, saito2024rgca}, our goal is a compact and light representation that does not need vast amounts of compute resources to generate novel expressions of the human.
Unfortunately, most of the methods show that to produce high-quality results, one needs to employ heavy CNN-based architectures which are not well suited for commodity devices and tend to slow down the rendering pipeline. Moreover, those models comprise dozens of millions of parameters creating heavy checkpoints that can easily exceed 500 MB. This ultimately creates a major issue for distributing and managing personalized models.
We tackle this problem by distilling a CNN-based architecture, leading to a personalized \textbf{\longname}, \textbf{\shortname} in short.
Our approach builds on Gaussian maps predicted from a modified UNet architecture \cite{Wang2023StyleAvatarRP} which is used for the UV space normalization required to build linear eigenbases.
Based on the per-subject trained CNN model, we bootstrap the \shortname by computing an ensemble of linear bases on the predicted Gaussian maps of the training frames.
The bases are refined on the training corpus using photometric losses while preserving their orthogonality.

These lightweight appearance bases are controlled with a relatively low number of parameters ranging from twenty up to fifty coefficients which can be specified w.r.t. the available compute resources and can for example be regressed by a ResNet-based model~\cite{Feng2020LearningAA}.
We demonstrate this for self-reenactment as well as cross-person animation, including a real-time demo in the suppl. video.

\medskip \noindent
In summary, our main contributions are:
\begin{enumerate}
    \item \longname (\shortname), a distillation technique of 3D Gaussian head avatar models built upon an ensemble of eigenbases.
    \item real-time (cross-person) animation of GEMs from single input images using a generalizable regressor.
\end{enumerate}

\section{Related Work}
\label{sec:related}
\begin{figure*}[ht!]
    \centering
    \includegraphics[width=1.0\textwidth]{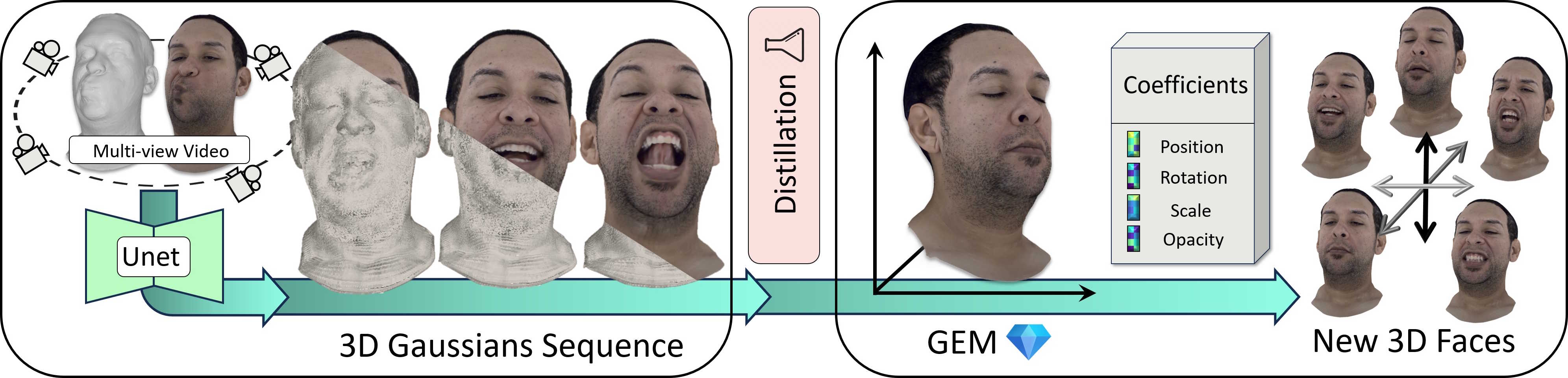}
    \caption{Given a multi-view video of a subject and mesh tracking, we create a dataset of 3D Gaussian point clouds for each frame in the sequence.
    Using this data, we distill a high-quality Gaussian Eigen Model (\shortname).
    \shortname is an ensemble of linear bases for each Gaussian primitive modality: position, opacity, scale, and rotation.
    Based on these bases, facial appearances are generated by a linear combination.}
    \label{fig:pipeline}
\end{figure*}

The majority of face representation and tracking techniques are based on parametric 3D morphable models (3DMM)~\cite{Blanz1999AMM,FLAME:SiggraphAsia2017}.
For a detailed overview, we refer to the state-of-the-art reports on face tracking and reconstruction~\cite{zollhoefer2018facestar}, the report on morphable models~\cite{egger2020morphablemodels}, and the two neural rendering state-of-the-art reports~\cite{tewari2020neuralrendering,tewari2022advances} that demonstrate how neural rendering can be leveraged for photo-realistic facial or full body avatars.
Next, we review the recent methods for photo-realistic 3D avatars generation which build appearance models using neural radiance fields (NeRF) \cite{Mildenhall2020NeRF} or volumetric primitives like 3D Gaussians \cite{Kerbl20233DGS}.

\subsection{NeRF-based avatars}

One of the first methods that combines a 3DMM and NeRF is NeRFace \cite{Gafni2020DynamicNR}, where a neural radiance field is directly conditioned by expression codes of the Basel Face Model (BFM)~\cite{Blanz1999AMM,Thies2016Face2FaceRF}.
This idea gave rise to many methods \cite{Zielonka2022InstantVH, Zheng2021IMA, grassal2022neural, Zheng2022PointAvatarDP, zhao2023havatar, xu2023latentavatar, Prinzler2022DINERDI, Gao2022ReconstructingPS} following a similar approach, but attaching the radiance fields more explicitly to the surface of the 3DMM, e.g., by using the 3DMM-defined deformation field.
For photorealistic results, some methods employ StyleGAN2-like architectures \cite{Karras2019AnalyzingAI} with a NeRF-based renderer~\cite{Kabadayi2023GANAvatarCP,An2023PanoHeadG3, Chan2021EfficientG3}.
Generative methods like EG3D~\cite{Chan2021EfficientG3} and PanoHead~\cite{An2023PanoHeadG3} employ GAN-based training to predict triplane features that span a NeRF.
GANAvatar~\cite{Kabadayi2023GANAvatarCP} applies this scheme to reconstruct a personalized avatar.

Close to our method is StyleAvatar~\cite{Wang2023StyleAvatarRP}.
Based on 3DMM tracking the method learns a personalized avatar that benefits from a StyleUNet which incorporates StyleGAN~\cite{Karras2019AnalyzingAI} to decode the final image.
Despite real-time capabilities, StyleAvatar suffers from artifacts produced by the image-to-image translation network that we explicitly avoid by using Gaussian maps which can compensate for tracking misalignments by predicting corrective fields for the 3D Gaussians.

\subsection{3D Avatars from Volumetric Primitives}

%
Using multiview images with a variational auto-encoder \cite{kingma2022autoencoding} and volumetric integration, Neural Volumes (NV) \cite{Lombardi2019NeuralV} encodes dynamic scenes into a volume which can be deformed by traversing a latent code $\textbf{z}$.
To better control the 3D space, Lombardi \etal \cite{Lombardi2021MixtureOV} introduce Mixture of Volumetric Primitives (MVP) a hybrid representation based on primitives attached to a tracked mesh which ultimately replaced the encoder from NV.
Each primitive is a volume represented as a small voxel with $32^3$ cells that store RGB and opacity values.
The final color is obtained by integrating values along a pixel ray.
This hybrid representation inspired many follow-up projects \cite{Cao2022AuthenticVA, Remelli2022DrivableVA, Li2023MEGANEME, Teotia2023HQ3DAvatarHQ, adaptiveshells2023}.
As an alternative to MVP primitives, 3D Gaussian Splatting (3DGS) \cite{Kerbl20233DGS} represents a volume as a set of anisotropic 3D Gaussians, which are equivalently described as ellipsoids, in contrast to isotropic spheres used in Pulsar \cite{Lassner2021PulsarES}.

Numerous methods \cite{Zielonka2025Drivable3D, Li2023AnimatableGL, saito2024rgca, Xu2023GaussianHA, zheng2023gpsgaussian, Qian2024gaussianavatars, feng2024gaussian, xie2024physgaussian, jiang2024vrgs, Pang2023ASHAG, kirschstein2024ggheadfastgeneralizable3d, giebenhain2024npganeuralparametricgaussian, zielonka2025synshot} capitalize on the speed and quality of 3DGS.
Qian \etal~\cite{Qian2024gaussianavatars} attach 3D Gaussians to the FLAME \cite{FLAME:SiggraphAsia2017} mesh surface and apply a deformation gradient similar to Zielonka et al. \cite{Zielonka2022InstantVH} to orient the Gaussians according to the local Frenet frames of the surface.
This method, however, does not utilize any information about expressions and, thus, struggles with pose-dependent changes (e.g., wrinkles, self-shadows) and, despite high-quality results, retrieves only a global static appearance model.
3D Gaussian blendshapes \cite{ma2024gaussianblendshapes} controls an avatar by linearly interpolating between optimized blendshapes using 3DMM expression coefficents. However, this method depends on an underlying 3DMM whereas GEM is a mesh-free representation.
Li \etal \cite{Li2023AnimatableGL} use a StyleUNet-like CNN architecture \cite{Wang2023StyleAvatarRP} to regress front and back Gaussian maps.
Employing a powerful CNN network on position maps, they achieve impressive results for human bodies with effects like pose-dependent wrinkle formation.
%

Please note that in this work, we focus on methods that directly output Gaussian primitives.
This is an important distinction from a branch of methods that follow Deferred Neural Rendering~\cite{Thies2019DeferredNR}, where a refinement CNN translates splatted features or coarse colors into the final image; for instance, Gaussian Head Avatars~\cite{Xu2023GaussianHA} and NGPA~\cite{giebenhain2024npganeuralparametricgaussian}.
This distinction is important because Gaussian primitives cannot be fully distilled into an eigenbasis in this context, as the refinement CNN network is required to complete the rendering directly in the image space.

\subsection{3DGS Compression Methods}

Recently, several methods \cite{Papantonakis_2024, Lee_2024_C3DGS, Lee_2024_CVPR, navaneet2023compact3d, fan2023lightgaussian, girish2024eaglesefficientaccelerated3d} have been proposed to reduce the memory footprint of 3D Gaussian Splatting (3DGS).
Papantonakis \etal \cite{Papantonakis_2024} apply codebook quantization to the Gaussian primitive properties, alongside pruning of Spherical Harmonic (SH) coefficients based on their final contribution.
In contrast to postprocessing approaches \cite{Papantonakis_2024, fan2023lightgaussian, Lee_2024_CVPR}, Compact3D \cite{navaneet2023compact3d} employs a single-stage process that jointly optimizes both the codebook entries and the primitives.
Fan \etal \cite{fan2023lightgaussian} calculate a significance score for each primitive by measuring its pixel hit count, thereby improving the pruning strategy.
Most of these methods target static scenes or time-conditioned environments, unlike our approach, which focuses on efficient, fully controllable head avatars. Nonetheless, these compression techniques could be adapted to our animatable avatars to reduce memory usage.

\section{Method}
\label{sec:main}
Recent dynamic 3D Gaussian Avatar methods show unprecedented quality, however, they require sophisticated and often compute-heavy CNN-based architectures \cite{Li2023AnimatableGL, Pang2023ASHAG, Xu2023GaussianHA} to capture high-frequency and dynamic details like pose-dependent wrinkles or self-shadows.
The aim of this paper is to build on top of this quality but remove the compute-intense architecture during inference.
Specifically, we propose to distill high-quality avatar models into lightweight linear animation models which we call GEMs.
A GEM is defined by an ensemble of eigenbases that span the space of the 3D Gaussian primitives.
These eigenbases are constructed via PCA applied on a dataset of per-frame Gaussian primitives, see \Cref{sec:gem}.
\textit{An important distinction compared to other neural avatars \cite{Zielonka2022InstantVH, Gafni2020DynamicNR, grassal2022neural, Lombardi2019NeuralV, Xu2023GaussianHA, Wang2023StyleAvatarRP} is that \shortname does not require a 3DMM \cite{FLAME:SiggraphAsia2017,bfm09} at test time. }
We demonstrate that a GEM can be directly driven by a monocular video using a generalized image-based regression network, see \Cref{sec:gem_driver}.

\subsection{Gaussian Eigen Model (GEM)}
\label{sec:gem}

For our distillation, we reconstruct a sequence of normalized Gaussian primitives $\pmb{D} = \{\pmb{G}_0, ..., \pmb{G}_{N-1}\}$.
As input, we assume a multi-view video of the subject with $N$ time frames.
Per time frame $i$, we reconstruct the 3D Gaussian pointcloud $\pmb{G}_i$, where $\pmb{G}_i$ contains the parameters that define the 3D Gaussians such as $\text{rotation } \vec{\theta}, \text{position } \vec{\phi}, \text{opacity } \vec{\alpha}, \text{scale } \vec{\sigma}$, and $\text{color } \vec{c}$ such that $\pmb{G}_i = \{\vec{\theta}, \vec{\phi}, \vec{\alpha}, \vec{\sigma}, \vec{c}\}$.

\paragraph{Reconstructing High-quality 3D Gaussian Primitives: }
We are following the idea of organizing the 3D Gaussians in 2D maps~\cite{Li2023AnimatableGL, Pang2023ASHAG, Xu2023GaussianHA, saito2024rgca}, where each pixel represents a 3D Gaussian with its parameters.
We propose an adapted CNN-architecture of Animatable Gaussians (AG)~\cite{Li2023AnimatableGL}, by merging the separate Style-U-Nets, reducing the convolutional layers, and operating in the UV space of the FLAME head model.
In addition, we are employing deformation gradients following Sumner \etal~\cite{Sumner2004DeformationTF} to handle the transformation from canonical to deformed space and treat the color as a global parameter.
We refer to the suppl. mat. for a detailed explanation of the architectural changes.
In comparison to the original AG model, our proposed CNN model produces slightly better results while being more efficient in terms of computing and memory.
Using this model, we generate the per-frame Gaussian primitives $\pmb{G}_i$ in the canonical space for all training time-frames.
Note that for this reconstruction, we follow Animatable Gaussians and, thus, FLAME-based tracking is required.
However, during inference, our model is independent of FLAME.

\paragraph{Distillation: }
Given $\pmb{D} = \{\pmb{G}_0, ..., \pmb{G}_{N-1}\}$, we build a personalized eigenbasis model, which is called \shortname.
We compute a statistical model for each Gaussian modality separately.
Specifically, we create individual bases for $\text{rotation } \pmb{B}_{\theta}, \text{position } \pmb{B}_{\phi}, \text{opacity } \pmb{B}_{\alpha}$, and $ \text{scale } \pmb{B}_{\sigma}$ with respective means $\vec{\mu}_\theta$, $\vec{\mu}_\phi$, $\vec{\mu}_\alpha$ and $\vec{\mu}_\sigma$ via Principle Complonent Analysis (PCA)~\cite{Jolliffe1986PrincipalCA}. 
Note that the color $\pmb{C}$ is optimized globally and, thus, acts as a classical texture without the need to apply PCA. To accurately learn dynamically moving Gaussians, we fixed the color to prevent it from dominating the image representation, otherwise, Gaussians could change their semantic meaning (e.g., a Gaussian could represent the lip in one state, and the teeth in the other deformation state).
Keeping the semantic meaning of specific Gaussians across deformation states is crucial for applying a PCA afterward.

\begin{figure}[t]
    \centering
    \includegraphics[width=\linewidth]{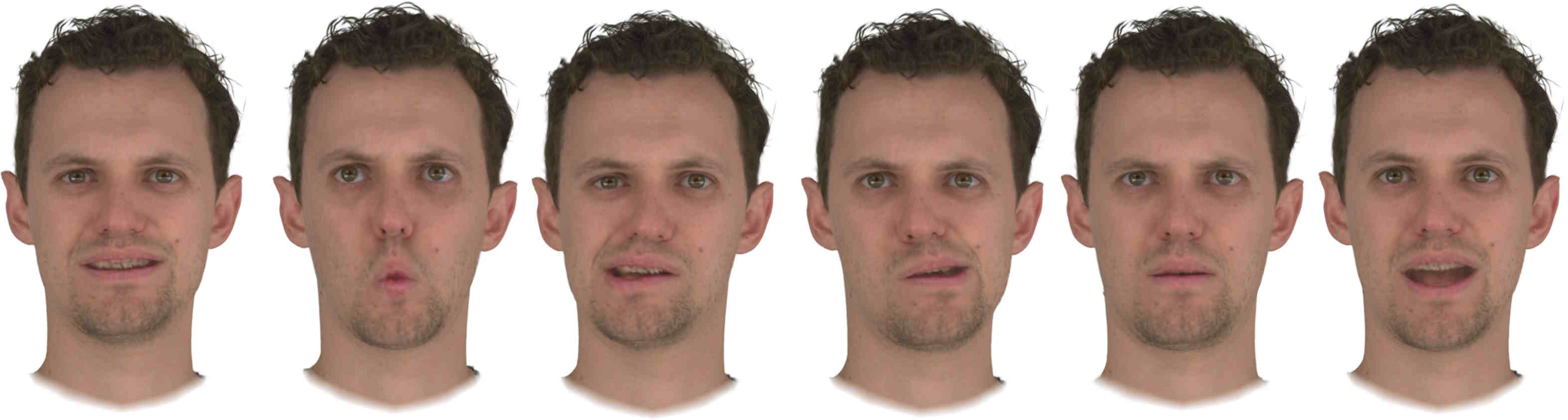}
    \caption{\textbf{Samples of a \shortname.} We display samples for the first three components of the position $\mathbf{k}_\phi$ eigenbasis of a GEM, showing diverse expressions. Note that GEM requires \textbf{no} parametric 3D face model like FLAME\cite{FLAME:SiggraphAsia2017}.}
    \label{fig:pca_traversal}
\end{figure}

A face model instance $\pmb{G}$ is represented as a linear combination of these bases:
\begin{equation}
\pmb{G} = \left\{ \vec{\mu}_i + \mathbf{B}_i \mathbf{k}_i \mid i \in \{\theta, \phi, \alpha, \sigma\} , \vec{c} \right\},
\label{eq:gem_forward}
\end{equation}

\noindent
where $\mathbf{k}_\theta$, $\mathbf{k}_\phi$, $\mathbf{k}_\theta$ and $\mathbf{k}_\sigma \in \mathbb{R}^M$ are the linear coefficients which are defining the facial expression state, assuming $M$ principal components.
As an example, \Cref{fig:pca_traversal} shows position parameter $\mathbf{k}_\phi$ sampled in the range of $[-3\sigma_\phi, 3\sigma_\phi]$ ($\sigma_\phi$ being the std. deviation).
As the Gaussian primitives $\mathbf{D}$ might contain tracking failures and misalignments, the principle components $\pmb{B}_{(\theta,\phi, \alpha, \sigma)}$ also contain artifacts as well.
We, therefore, refine the bases using the training images directly, by applying a photometric reconstruction loss.
We employ the same objectives from the CNN model training (see supp. mat).
\begin{equation}
\label{formula: color loss}
\mathcal{L}_{Color} = 
        (1 - \omega) \mathcal{L}_1 + \omega\mathcal{L}_{\text{D-SSIM}} + \zeta\mathcal{L}_{\text{VGG}}
\end{equation}
We refine the base vectors for around 30k iterations.
To ensure that the individual bases stay orthonormal throughout this refinement, every 1k steps, we orthogonalize the bases using QR decomposition.
This refinement improves the training PSNR errors from 34.75dB to 36.68dB and 36.85dB for the training steps 0k, 5k, and 30k, respectively.
Throughout our experiments, we did not encounter overfitting issues with this scheme.
The reconstruction metrics on two randomly selected test sequences with refinement are: PSNR: \textbf{31.51}, LPIPS: \textbf{0.091}, SSIM: \textbf{0.936}; and without: PSNR: 31.38, LPIPS: 0.094, SSIM: 0.933.

\begin{figure}[t]
    \centering
    \includegraphics[width=1.0\columnwidth]{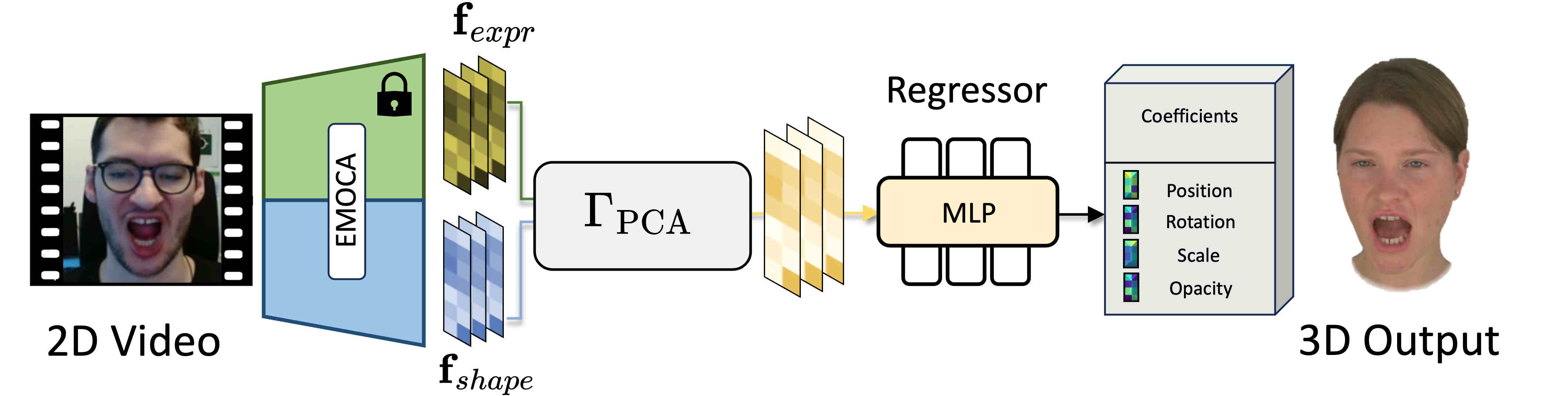}
    \caption{\textbf{Image-based animation.} One of the applications of our \shortname is real-time (cross)-reenactment. For that, we utilize generalized features from EMOCA~\cite{Danek2022EMOCAED} and build a pipeline to regress the PCA coefficients of our model from an input image/video.}
    \label{fig:regressor_arch}
\end{figure}

\begin{figure*}[h]
    \centering
    \setlength{\unitlength}{0.1\textwidth}
    \begin{picture}(10, 3.5)
    \put(0, 0){\includegraphics[width=\textwidth, clip, trim=0cm 0cm 0cm 22cm]{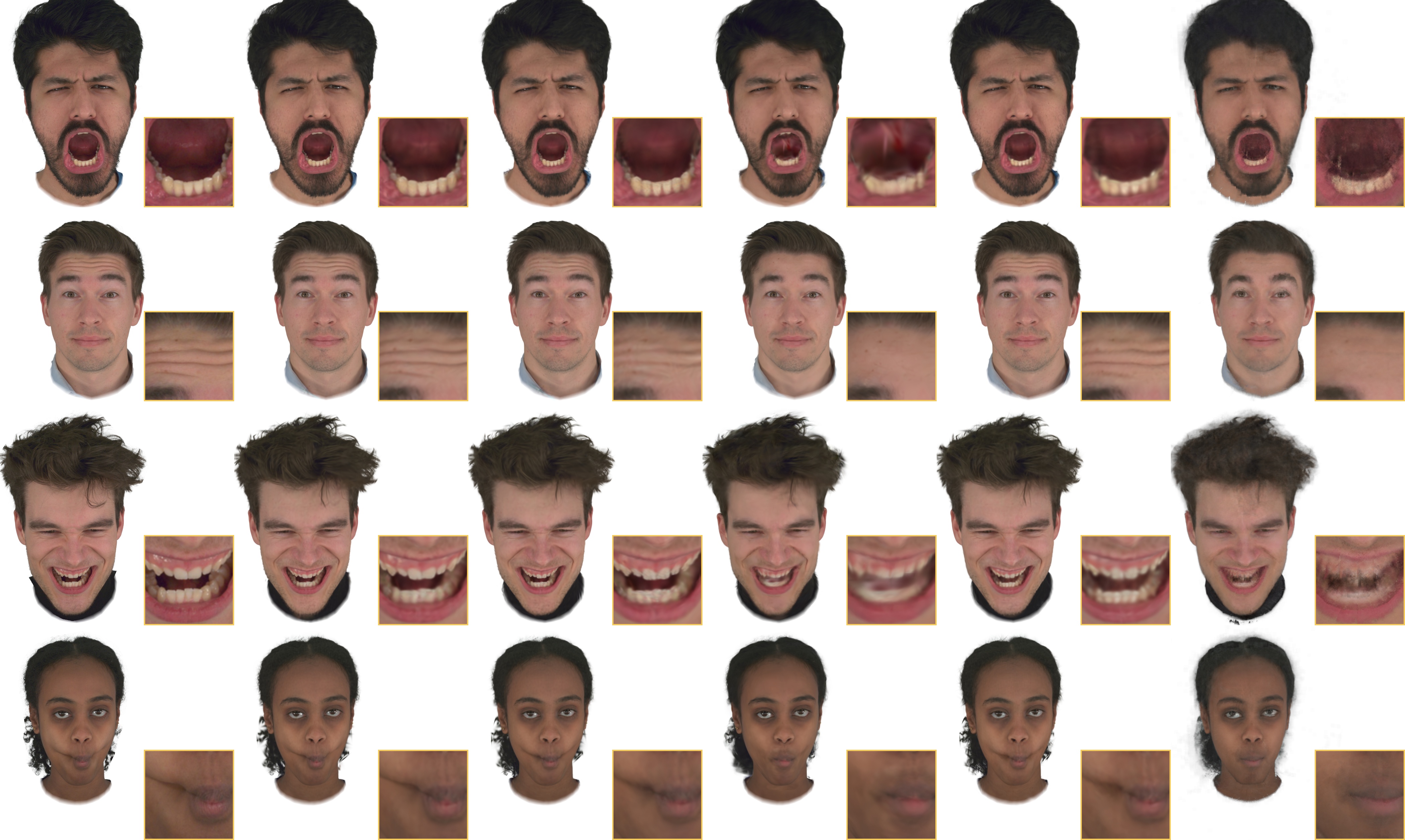}}
    \put(0, -0.4){Ground Truth}
    \put(2.0, -0.4){Ours GEM}
    \put(3.65, -0.4){Ours Net}
    \put(5.5, -0.4){GA~\cite{Qian2024gaussianavatars}}
    \put(7.1, -0.4){AG~\cite{Li2023AnimatableGL}}
    \put(8.6, -0.4){INSTA~\cite{Zielonka2022InstantVH}}
    \end{picture}
    \vspace{0.25cm}
    \caption{\textbf{Novel view synthesis.} Both, our CNN and GEM show better performance on novel views, especially, in the region of the mouth interior and wrinkles. In this experiment, we are following the evaluation of Gaussian Avatars~\cite{Qian2024gaussianavatars} and demonstrate novel viewpoint generation. GEM is obtained throughout analysis-by-synthesis fitting \cite{Blanz1999AMM, Thies2016Face2FaceRF}. Note that the expressions are seen during training.}
    \label{fig:novel_view}
\end{figure*}

\subsection{Image-based Animation}
\label{sec:gem_driver}

Expressions for a \shortname are fully defined by their coefficients $\mathbf{k}_\theta$, $\mathbf{k}_\phi$, $\mathbf{k}_\theta$ and $\mathbf{k}_\sigma$.
This is a similar idea to codec avatars~\cite{ma2021pixel}, however, our approach does not need additional pixel shaders in the form of a small regression MLP.
There are several ways to obtain the coefficients of a GEM, for example, one can employ analysis-by-synthesis-based optimization or regression.
Analysis-by-synthesis \cite{Blanz1999AMM} is the backbone of current avatar methods, as they use photometric or depth-based face trackers to sequentially optimize the coefficients of the underlying 3DMMs like FLAME \cite{Thies2016Face2FaceRF, Zielonka2022InstantVH, grassal2022neural} which is typically slow.
As a fast, but more imprecise alternative, regressors like DECA \cite{Feng2020LearningAA} or EMOCA \cite{Danek2022EMOCAED} can be used which are built on a ResNet backbone and regress FLAME parameters directly from an image.
We apply several modifications to the EMOCA model, see \ref{fig:regressor_arch}.
We use intermediate features of the pre-trained EMOCA network denoted as $\Theta(\mathbf{I}_i)$ where $\mathbf{I}_i$ is the current image. 
EMOCA's architecture comprises two ResNet networks; one to extract expression features $\mathbf{f}_{expr} \in \mathbb{R}^{2048}$ and the second for shape $\mathbf{f}_{shape} \in \mathbb{R}^{2048}$, both are followed by final MLPs to regress corresponding FLAME parameters.
As we do not rely on FLAME, we remove the last hidden layer of the final MLP obtaining two feature vectors which we combine into one $\mathbf{f} \in \mathbb{R}^{2 \times 1024}$ vector.
For these features, we build a PCA layer with a basis denoted as $\mathbf{\hat{R}}$ using the training frames from five frontal cameras of NeRSemble.
Note that we use relative features $\mathbf{r} = \mathbf{f} - \mathbf{f}_{neutral}$ in this PCA layer. 
The neutral reference frame $\mathbf{f}_{neutral} = \Theta(\mathbf{I}_{neutral})$ to compute these relative features is selected manually from the video, similar to Face2Face~\cite{Thies2016Face2FaceRF}.
During training, for each frame, we project $\mathbf{r}$ onto the PCA manifold using the first $50$ principal components to restrict and regularize training. 
Finally, we use their corresponding PCA coefficients:
\begin{equation}
\begin{split}
\label{formula: residual projections}
\mathbf{\kappa} &= (\mathbf{r} - \mathbf{\bar{R}})\mathbf{\hat{R}}^T, \\
\end{split}
\end{equation}
where $\mathbf{\bar{R}}$ is the relative PCA model mean.
The projected coefficients are passed through a small MLP that produces a vector of \shortname coefficients \mbox{$\mathbf{k} = \{
\mathbf{k}_\theta, \mathbf{k}_\phi, \mathbf{k}_\theta, \mathbf{k}_\sigma
\}$}:
\begin{equation}
\begin{split}
\label{formula: regressor model}
\mathbf{k} &= 3 \cdot \sigma_k \cdot\tanh(\textrm{MLP}(\mathbf{\kappa})).
\end{split}
\end{equation}
The MLP has three hidden layers with 256 neurons each and ReLU activations.
We use a scaled $\tanh$ activation function for the output to restrict the prediction to be in $[-3 \cdot \sigma_k, 3 \cdot \sigma_k]$, $\sigma_k$ being the respective standard deviation of the coefficients $\mathbf{k}$, obtained from the PCA.
The final primitives are obtained by Eq. \ref{eq:gem_forward} and splatted using 3DGS.

\section{Results}
\label{sec:results}

\begin{figure*}[h]
    \centering
    \setlength{\unitlength}{0.1\textwidth}
    \begin{picture}(10, 4)
    \put(0, 0){\includegraphics[width=\textwidth, clip, trim=0cm 0cm 0cm 20cm]{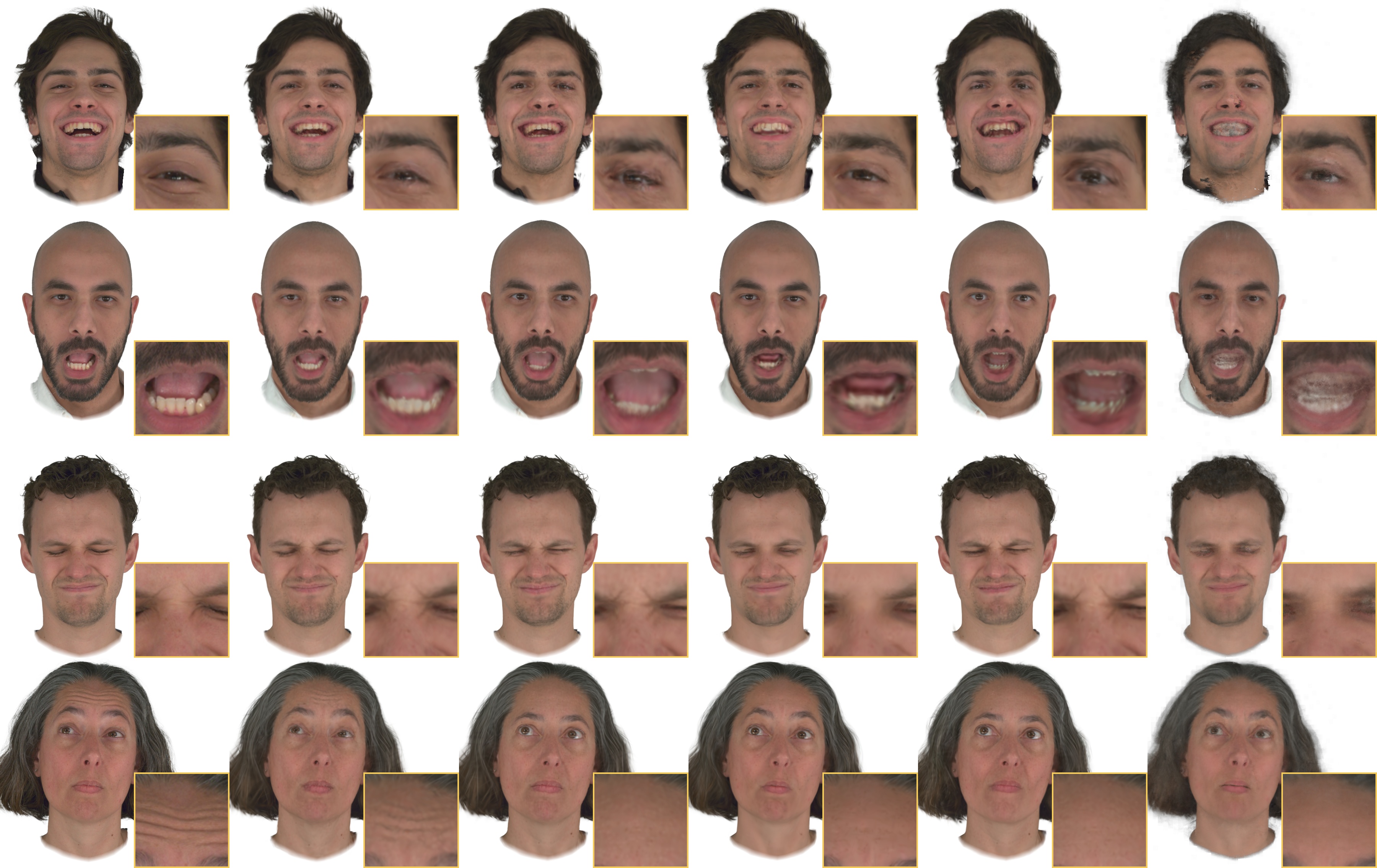}}
    \put(0, -0.4){Ground Truth}
    \put(2.0, -0.4){Ours GEM}
    \put(3.65, -0.4){Ours Net}
    \put(5.5, -0.4){GA~\cite{Qian2024gaussianavatars}}
    \put(7.1, -0.4){AG~\cite{Li2023AnimatableGL}}
    \put(8.6, -0.4){INSTA~\cite{Zielonka2022InstantVH}}
    \end{picture}
    \vspace{0.25cm}
    \caption{\textbf{Novel view and expression synthesis.} Our \longname shows better results in regions like teeth, wrinkles, and self-shadows compared to other methods that struggle with artifacts.
    }
    \label{fig:test}
\end{figure*}

We evaluate GEM on the NeRSemble~\cite{Kirschstein2023NeRSembleMR}, where tracked meshes~\cite{Qian2024gaussianavatars} and synchronized images from 16 cameras with a resolution of $802\times550$ are available.
Our baselines are Gaussian Avatars (\textbf{GA})~\cite{Qian2024gaussianavatars} which is neural network-free (Gaussians are attached to the FLAME model), our implementation of Animatable Gaussians (\textbf{AG}) ~\cite{Li2023AnimatableGL} which is based on CNN-predicting Gaussian maps, and INSTA~\cite{Zielonka2022InstantVH} which uses dynamic NeRF~\cite{Mildenhall2020NeRF}.
Note that all baselines require at least two stages: (i) construct the avatar, and (ii) get the parameters to drive it.
Most of them use offline tracking with additional objectives like hair reconstruction \cite{Qian2024gaussianavatars, giebenhain2024mononphm}, which does not work for real-time applications despite the avatar model's rendering being real-time.
Importantly, in our approach, we introduce a third step, i.e., the construction of the eigenbasis (GEM), which only introduces \textbf{negligible} computational costs \textbf{($\sim$ 1 min)} in comparison to the avatar reconstruction itself.
For the comparison, we present both of our appearance models, the StyleUNet-based architecture (\textbf{Ours Net}) and the distilled linear Eigen model (\textbf{Ours GEM}) which we evaluate using analysis-by-synthesis fitting to the target images following \cite{Zielonka2022TowardsMR, Thies2016Face2FaceRF, Blanz1999AMM}.
Additionally, we present cross-reenactment results based on our coefficient regressor, compared to the baselines that use FLAME meshes regressed by DECA \cite{Feng2020LearningAA}.
Relative expression transfer based on ground truth meshes~\cite{Qian2024gaussianavatars} can be found in the supp. mat.
All of the methods are evaluated using several image space metrics on novel expressions and novel views, following the test and novel-view split of Qian \etal~\cite{Qian2024gaussianavatars}.
For our GEM models, we use 50 components distilled from $256^2$ textures which give around 60k active Gaussians.
Animatable Gaussians~\cite{Li2023AnimatableGL} uses a similar amount of primitives for front and back textures and Gaussian Avatars~\cite{Qian2024gaussianavatars} around 100k Gaussians.
\begin{table}[t]
    \centering
    \footnotesize
        \begin{tabular}{l|rrrr}
        Method & PSNR $\uparrow$ & LPIPS $\downarrow$ & SSIM $\uparrow$ & L1 $\downarrow$ \\
        \midrule       
        AG \cite{Li2023AnimatableGL}   & 32.4166 & 0.0712 & 0.9614 & 0.0066 \\
        GA \cite{Qian2024gaussianavatars}  & 31.3197 & 0.0786 & 0.9567 & 0.0075 \\
        INSTA \cite{Zielonka2022InstantVH}   & 27.7786 & 0.1232 & 0.9294 & 0.0163 \\
        \bottomrule
        Ours Net                             & 32.4622 & 0.0713 & 0.9617 & 0.0067 \\
        Ours GEM                             & \textbf{33.5528} & \textbf{0.0678} & \textbf{0.9662} & \textbf{0.0061} \\
        \bottomrule
        \end{tabular}
    \caption{\textbf{Novel viewpoint evaluation} is conducted on a withhold camera from the 16 cameras used for training. Note that the expression has been seen during training, and only the view is new.}
    \label{tab:color_errors_novel}
\end{table}

\begin{table}[t]
  \centering
  \footnotesize
    \begin{tabular}{l|rrrr}
    Method & PSNR $\uparrow$ & LPIPS $\downarrow$ & SSIM $\uparrow$ & L1 $\downarrow$ \\
    \midrule
    
    AG \cite{Li2023AnimatableGL}  & 29.0114 & 0.0812 & 0.9429 & 0.0099 \\
    GA \cite{Qian2024gaussianavatars}  & 28.3137 & 0.0815 & 0.9433 & 0.0102 \\  
    INSTA \cite{Zielonka2022InstantVH}     & 27.9181 & 0.1153 & 0.9340 & 0.0128 \\
    \bottomrule
    Ours Net                               & 29.2454 & 0.0777 & 0.9448 & 0.0096 \\
    Ours GEM                               & \textbf{32.6781} & \textbf{0.0675} & \textbf{0.9633} & \textbf{0.0069} \\
    \bottomrule
    \end{tabular}
  \caption{\textbf{Evaluation on novel expressions} and views show improved results of GEM optimized using analysis-by-synthesis compared to others. \Cref{fig:test} shows the corr. qualitative results.}
  \label{tab:color_errors_test}
\end{table}

\subsection{Image Quality Evaluation}
To evaluate our method, we measure the color error in the image space using the following metrics: PSNR (dB), LPIPS \cite{zhang2018unreasonable}, L1 loss, and structural similarity (SSIM). We follow the evaluation scheme from Gaussian Avatars \cite{Qian2024gaussianavatars}, using their train and validation split.
The evaluation of \shortname was generated by sequentially fitting the coefficients to each image using photometric objectives.
Note that the baselines use the FLAME model with offsets for the tracking, while \shortname can directly be used for tracking.

Table \ref{tab:color_errors_test} presents results on novel expressions evaluated on all 16 cameras. Both the quantitative and qualitative results depicted in Figure \ref{fig:test} show that our PCA model produces  
fewer artifacts, especially for regions like teeth or facial wrinkles. 
Table \ref{tab:color_errors_novel} contains an evaluation where we measure errors on novel viewpoints.
The results demonstrate that our CNN-based appearance model outperforms other neural methods, while our linear eigenbasis \shortname achieves the highest quality. This is due to the 'direct' analysis-by-synthesis approach, which fully leverages the expressiveness and detail of our photorealistic appearance model, without the limitations imposed by 3DMMs such as FLAME.
Moreover, Figure \ref{fig:novel_view} shows qualitative results of our method on novel views.
As can be seen, we better capture high-frequency details, pose-dependent wrinkles, and self-shadows  - something which is not possible for methods like Gaussian Avatars \cite{Qian2024gaussianavatars} or INSTA \cite{Zielonka2022InstantVH}, since they either do not use expression-dependent neural networks or limit the conditioning to a small region only.

\begin{table}[b]
    \centering
    \resizebox{0.9\linewidth}{!}{
        \begin{tabular}{l|rrrr}
        Method & $\textbf{E}_{feat}$cos $\uparrow$ & $\textbf{E}_{feat}$ $\mathcal{L}_1$ $\downarrow$ & FID $\downarrow$ & FPS $\uparrow$ \\
        \midrule
        AG       & 0.9396 & 5.3399 & 0.4093 & 16.51  \\
        GA       & 0.8917 & 6.6141 & 0.5593 & 142.71 \\
        INSTA    & 0.9087 & 6.3153 & 0.5299 & 20.62  \\
        Ours Net & 0.9440 & 5.1044 & 0.3685 & 35.77  \\
        Ours GEM & 0.9381 & 5.3197 & 0.4286 & 201.70 \\
        \bottomrule
        \end{tabular}
    }
    \caption{\textbf{Cross-reenactment evaluation} employing EmoNet features and FID score.}
    \label{tab:fid}
\end{table}

\begin{figure*}[t]
    \centering
    \setlength{\unitlength}{0.1\textwidth}
    \begin{picture}(10, 4)
    \put(0, 0){\includegraphics[width=\textwidth]{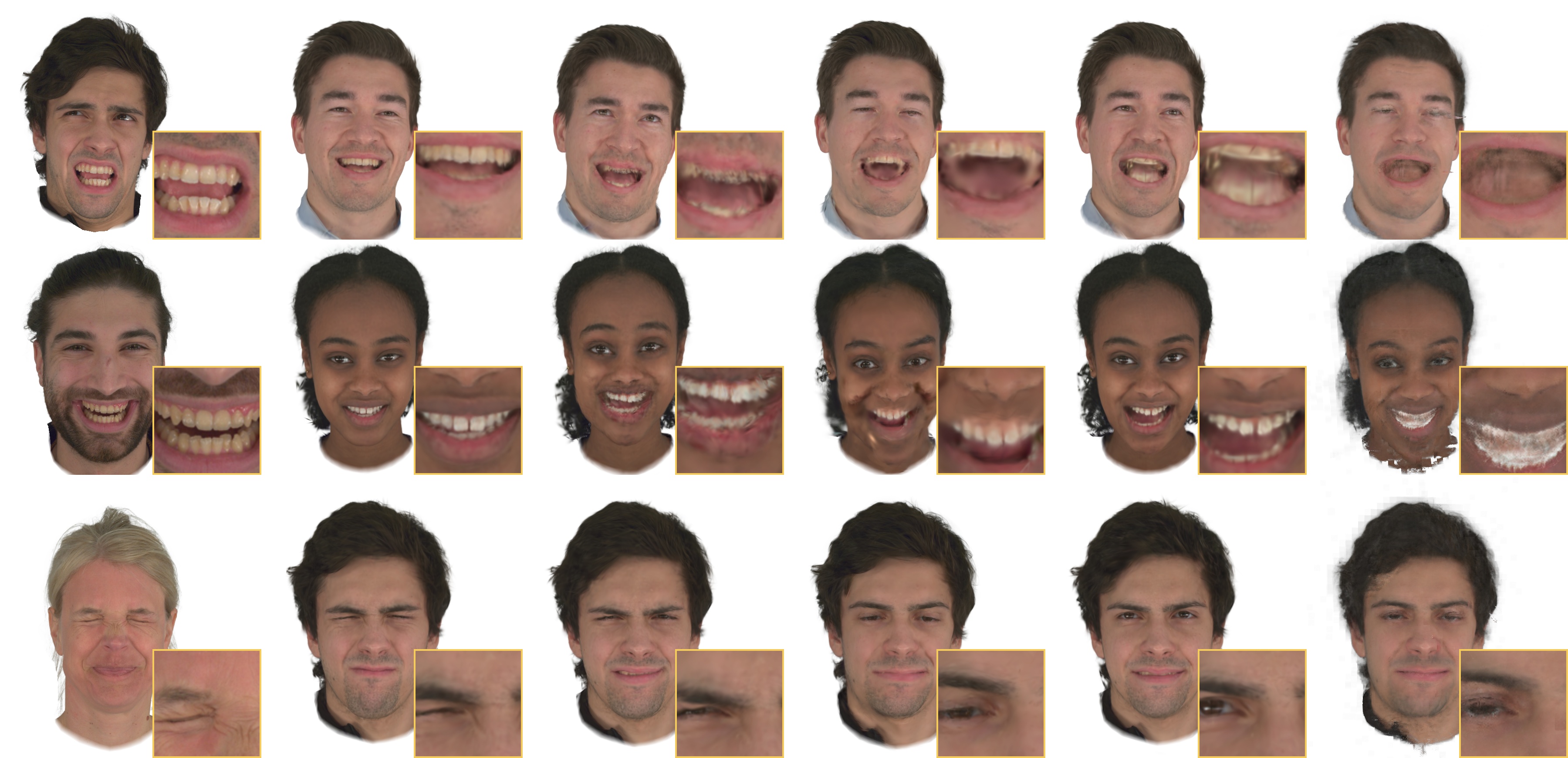}}
    \put(0, -0.4){Source Actor}
    \put(2.0, -0.4){Ours GEM}
    \put(3.65, -0.4){Ours Net}
    \put(5.5, -0.4){GA~\cite{Qian2024gaussianavatars}}
    \put(7.1, -0.4){AG~\cite{Li2023AnimatableGL}}
    \put(8.6, -0.4){INSTA~\cite{Zielonka2022InstantVH}}
    \end{picture}
    \vspace{0.2cm}
    \caption{\textbf{Facial cross-person reenactment using an image-based regressor.} The reenactment of the baselines is performed using relative transfer between FLAME meshes regressed by EMOCA compared to our \shortname regressor network (Ours GEM).}
    \label{fig:transfer_deca}
\end{figure*}

\subsection{Cross-person Reenactment Evaluation}
Facial cross-person reenactment transfers expressions from the source actor to the target actor.
For this, the baseline methods require tracked meshes obtained by fitting the 3DMM model for each frame of the source actor sequence.
As an alternative to optimization-based tracking, a (monocular) regressor like EMOCA \cite{Danek2022EMOCAED}, can predict such tracked meshes in real-time.
We demonstrate this in \Cref{fig:transfer_deca}, where GEM is driven by our image-based regressor and the others by EMOCA.
As shown, our network-based method and \shortname produce sharp results, while the baseline methods struggle to extrapolate to new expressions, displaying severe artifacts in appearance.
Our approach effectively regularizes the regressed coefficients, ensuring that the predicted avatar remains in the training distribution and thereby avoids artifacts seen in INSTA or Gaussian Avatars.
\begin{figure*}[ht!]
    \centering
    \centering
    \setlength{\unitlength}{0.1\textwidth}
    \begin{picture}(10, 1.3)
    \put(0, 0){\includegraphics[width=1.0\textwidth]{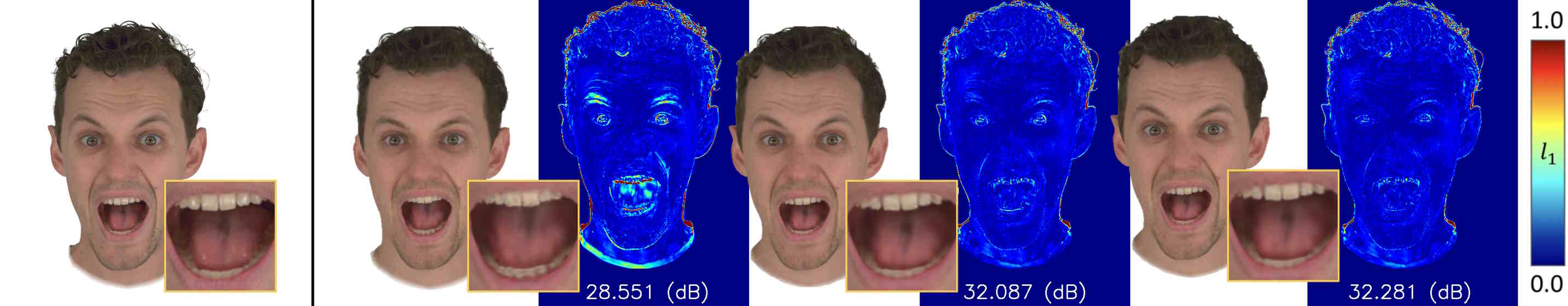}}
    \put(0.2, -0.2){Ground Truth}
    \put(3.1, -0.2){\#10}
    \put(5.4, -0.2){\#30}
    \put(7.7, -0.2){\#50}
    \end{picture}
    \caption{\textbf{Compression error depending on the number of used principal components in GEM.} The heatmaps show the photometric $\ell_1$-error for 10, 30, and 50 components using $128^2$ Gaussian maps. See suppl. doc. for additional evaluations.}
    \label{fig:pca_heatmap}
\end{figure*}
Drawing inspiration from EMOCA \cite{Danek2022EMOCAED}, we further assess cross-re-enactment quantitatively by leveraging emotion recognition feature vectors from both the source image and the resulting cross-re-enactment, utilizing EmoNet \cite{toisoul2021estimation}.
For each pair of input and output images, we predict EmoNet features and measure cosine distance and $\mathcal{L}_1$ error between them. We report the numbers in the \Cref{tab:fid}.
Additionally, we also report FID scores \cite{heusel2018ganstrainedtimescaleupdate} and rendering speed.
Our method achieves on-par quality with the CNN-based solution while maintaining the highest frame rates and outperforming GA in terms of quality. 
\subsection{GEM Ablation Studies}
\begin{figure}[h]
    \centering
    \footnotesize
    \setlength{\unitlength}{0.1\columnwidth}
    \begin{picture}(10, 2)
    \put(-0.2, 0){\includegraphics[width=1.0\columnwidth]{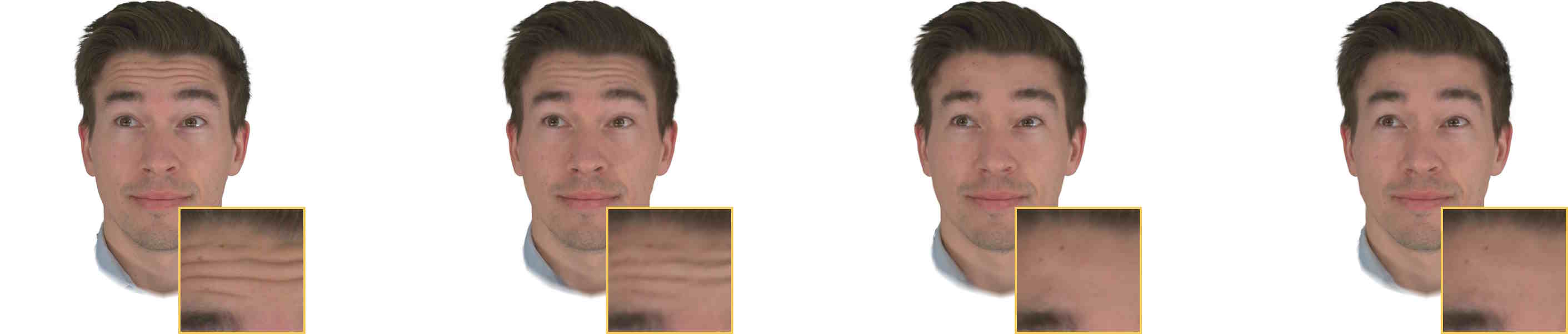}}
    \put(0.0, -0.4){Ground Truth}
    \put(2.6, -0.4){Ours \#30 $128^2$}
    \put(5.4, -0.4){GA \cite{Qian2024gaussianavatars} $128^2$}
    \put(8.0, -0.4){GA \cite{Qian2024gaussianavatars} $256^2$}
    \end{picture}
    \caption{The quality comparison to Gaussian Avatars. Note that we do \textbf{not} need an additional FLAME model which weighs 90MB.}
    \label{fig:ga_avatar_size_comparison}
\end{figure}

We are interested in the compression error introduced by the projection on different amounts of principal components used in GEM, also concerning the memory consumption.
\begin{figure}[h]
    \centering
    \includegraphics[width=1.0\linewidth]{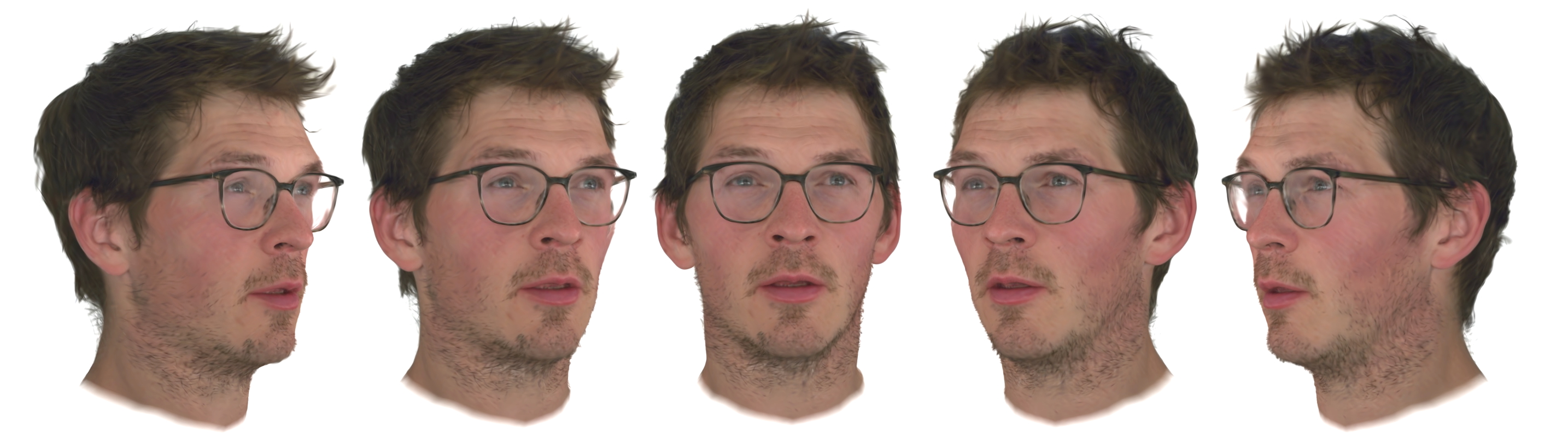}
    \caption{Despite fixed topology and predefined texture size GEM faithfully represents facial attributes like glasses.
    }
    \label{fig:glasses}
\end{figure}
Our smallest model weighs as little as \textbf{7MB} using only 10 components of the eigenbasis.
This is almost \textbf{12} times less than our smallest CNN-based model and almost \textbf{70} times less than Animatable Gaussians \cite{Li2023AnimatableGL}.
In contrast to neural networks, we can easily trade quality over size which is very useful in the context of different commodity devices with reduced compute capabilities.
Table \ref{tab:gem_ablation_size} presents how compression affects the quality of reconstruction, where we evaluate a sequence with $\sim1$k frames for a single actor under a novel view.
As expected, using only 10 components impacts the quality the most, however, the results are still of high quality, see \Cref{fig:pca_heatmap}.
Gaussian Avatars \cite{Qian2024gaussianavatars} offers a small size of the stored Gaussians cloud, ranging from 5MB, and 14MB without the FLAME model for $128^2$, $256^2$ Gaussians, respectively.
However, the quality of reconstruction lacks wrinkle details and sharpness as can be seen in \Cref{fig:ga_avatar_size_comparison}.
In comparison to Gaussian Avatars \cite{Qian2024gaussianavatars}, our model does not require FLAME during inference which is an additional \textbf{90MB}.
\begin{table}[h]
    \centering
    \label{tab:pca_size_mb}
    \resizebox{1.0\linewidth}{!}{
        \begin{tabular}{c|ccc|ccc|ccc}
        \toprule
        \multirow{2}{*}{\#Comp} &
        \multicolumn{3}{c}{$128^2$} &
        \multicolumn{3}{c}{$256^2$} &
        \multicolumn{3}{c}{$512^2$} \\
        \cline{2-10}
            & PSNR $\uparrow$ & Size MB & FPS $\uparrow$ & PSNR $\uparrow$ & Size MB & FPS $\uparrow$ & PSNR $\uparrow$ & Size MB & FPS $\uparrow$ \\
            \midrule
            10 & 31.81 & 7  & 237.96 & 31.88 & 28  & 210.03 & 32.23 & 113 & 130.46  \\
            30 & 34.20 & 20 & 241.31 & 34.17 & 83  & 208.19 & 34.84 & 333 & 112.73  \\
            50 & 34.67 & 34 & 238.7  & 34.61 & 138 & 201.70 & 35.45 & 553 & 117.45  \\
            \hline
            Ours Net   & 33.97 & 82 & 47.70  & 34.99 & 109 & 35.77 & 35.02 & 178 & 26.31  \\
            AG \cite{Li2023AnimatableGL} & 33.77 & 487 & 18.93 & 34.40 & 529 & 16.51 & 35.15 & 636 & 13.08  \\
            \bottomrule
            \end{tabular}
    }
    \caption{\textbf{Ablation of GEM.} Even with 10 principle components a high PSNR of $31.81dB$ is achieved, while taking only 7MB of memory. In contrast to fixed-sized neural networks, the \shortname can be adjusted on the fly depending on the hardware. Moreover, since evaluation requires a single dot product for forward pass the rendering speed is around four times higher than our network. The speed evaluation was done using a single Nvidia A100 GPU.}
    \label{tab:gem_ablation_size}
\end{table}
Figure~\ref{fig:glasses} demonstrates that our method is able to handle different topologies (subject wearing glasses), despite utilizing a fixed UV space.

\section{Discussion}
\label{sec:conclusion}
We design a universal method capable of distilling 3DGS-based avatar solutions into a lightweight representation, GEM, provided that normalized input across training frames is available.
The only requirement to successfully distill GEM is to have a dataset with Gaussian-image pairs across the training sequences.
Our results show that a compact representation of the linear basis produces state-of-the-art results in terms of quality and speed.
Note that to achieve wrinkle-level details, the generator itself has to produce high-quality outputs.
Our distillation technique can be applied to existing methods like \cite{zielonka2025synshot}, making them lightweight and compact.
GEM is well-suited for commodity devices, generating Gaussian primitives by a simple linear combination of the basis vectors.
This potential has promising implications for tasks like holoportation, audio-driven avatars, and virtual reality.

\paragraph{Limitations}: The PCA-based GEM models have a global extent which is useful for some applications, but it also means that we cannot control local changes and produce more combinations of local features.
Thus, further work could include incorporating a localized PCA basis~\cite{Neumann2013SparseLD} for better avatar control, which could potentially enable a wider range of expressions outside the training set.
Other limitations are; side-view generalization which results in unstable expressions and personalization. For new subjects a new representation has to be learned from multi-view data.
An interesting future avenue is to create a statistical model across subjects.

\section{Conclusion}
We have proposed \longname, a linear appearance model that represents photo-realistic head avatars.
The simplicity of this appearance model results in massively reduced compute requirements in comparison to CNN-based avatar methods. Although the idea is simple, it offers many interesting downstream applications. The lightweight representation could improve the management, sharing, and applicability of avatars.
Moreover, GEM simplifies the process of online avatar animation from RGB images and increases flexibility by balancing memory and quality trade-offs through additional control over the number of eigenbases.
Our distillation approach can be applied to existing methods, making them available for compression.
We demonstrate how GEMs can be used in scenarios like self-reenactment and cross-person animation, even in real-time.

\smallskip
\paragraph{Acknowledgements}
The authors thank the International Max Planck Research School for Intelligent Systems (IMPRS-IS) for supporting WZ. JT was supported by the ERC Starting Grant LeMo (101162081). All the data were processed outside Google.

{
    \small
    \bibliographystyle{ieeenat_fullname}
    \bibliography{main}
}

\clearpage
\appendix
\twocolumn[{%
\renewcommand\twocolumn[1][]{#1}%
\begin{center}
    \textbf{\Large{\longname \\ -- Supplemental Document --}}
    \centering
    \vspace{0.5cm}
    \captionsetup{type=figure}
    \includegraphics[width=1.0\textwidth]{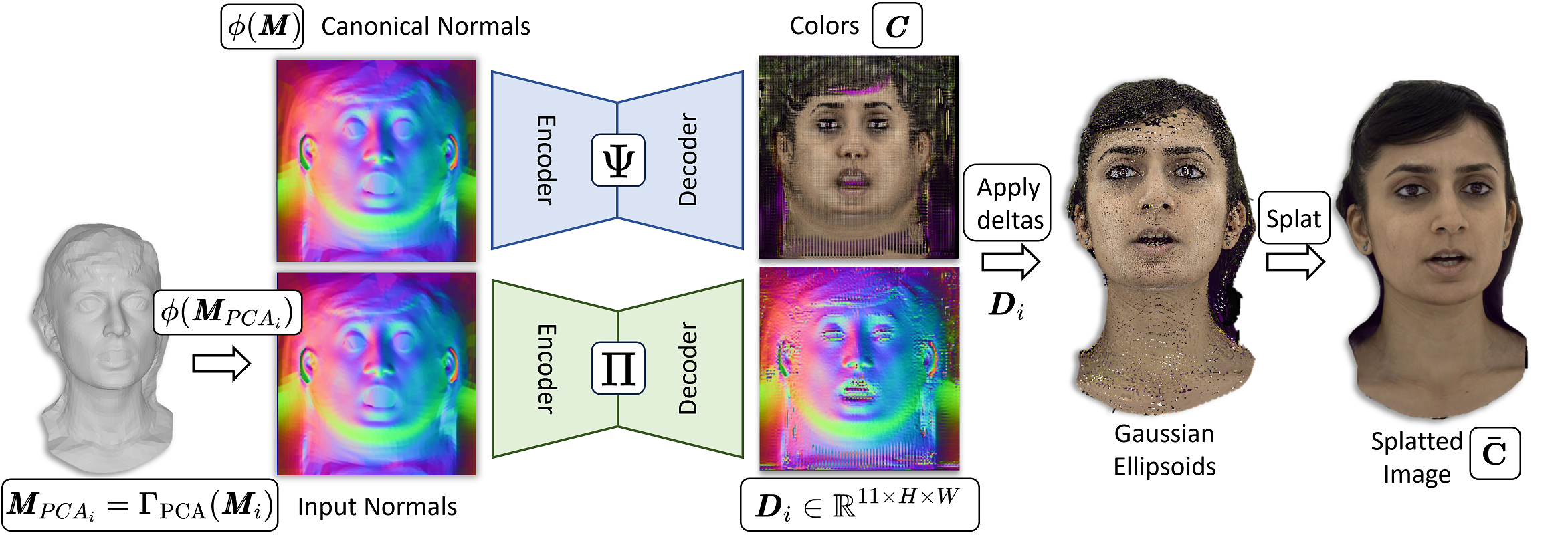}
    \caption{\textbf{CNN-based Gaussian Avatar Pipeline.} Our CNN model produces delta Gaussian maps $\pmb{D}_i$~\cite{Li2023AnimatableGL, Pang2023ASHAG} and static color $\pmb{C}$ from a multi-view video. Similarly to Animatable Gaussians~\cite{Li2023AnimatableGL}, we constrain the network to operate in a reduced linear space, i.e., the per-frame mesh $\pmb{M}_i$ is projected on a PCA basis \mbox{$\pmb{M}_{PCA_i}=\Gamma_{\text{PCA}}(\pmb{M}_i)$} which is then input to the network $\Pi$ after converting the mesh to a normal map $\phi(\pmb{M}_{PCA_i})$. The static color network is conditioned on neutral mesh $\pmb{M}$.
    }
    \label{fig:supp-teaser}
\end{center}%
}]

\section{3D Gaussian Splatting Preliminaries}
3D Gaussian Splatting (3DGS)~\cite{Kerbl20233DGS} is an alternative approach to Neural Radiance Field (NeRF)~\cite{Mildenhall2020NeRF} for static multi-view scene reconstruction and rendering under novel view.
Kerbl et al.~\cite{Kerbl20233DGS} parameterize the space as scaled 3D Gaussians~\cite{Wang2019DifferentiableSS, Kopanas2021PointBasedNR} with a 3D covariance matrix $\mathbf{\Sigma}$ and mean $\mathbf{\mu}$:
\begin{equation}
\label{formula: gaussian's formula}
    G(\mathbf{x})=e^{-\frac{1}{2}(\mathbf{x-\mu})^T\mathbf{\Sigma}^{-1}(\mathbf{x-\mu})}.
\end{equation}
To render this representation, Zwicker et al.~\cite{Zwicker2001SurfaceS} employ the projection of 3D Gaussians onto the image plane using the formula $\mathbf{\Sigma}^{\prime} = \mathbf{A}\mathbf{W}\mathbf{\Sigma} \mathbf{W}^T\mathbf{A}^T$, where $\mathbf{\Sigma}^{\prime}$ represents the covariance matrix in 2D space. Here, $\mathbf{W}$ denotes the view transformation, and $\mathbf{A}$ represents the projective transformation.
To avoid direct optimization of the covariance matrix $\mathbf{\Sigma}$ which must be positive semidefinite, Kerbl et al. \cite{Kerbl20233DGS} use scale $\mathbf{S}$ and rotation $\mathbf{R}$ which equivalently describes 3D Gaussian as a 3D ellipsoid $ \mathbf{\Sigma} = \mathbf{R}\mathbf{S}\mathbf{S}^T\mathbf{R}^T$.
Finally, 3DGS follows Ramamoorthi et al. \cite{Ramamoorthi2001AnER} 
to approximate the diffuse part of the BRDF \cite{Goral1984ModelingTI} as spherical harmonics (SH) to model global illumination and view-dependent color.
Four bands of SH are used which results in a 48 elements vector.

\section{Appearance Maps Generator}
Similarly to StyleAvatar~\cite{Wang2023StyleAvatarRP} and Animatable Gaussians \cite{Li2023AnimatableGL}, we use a StyleGAN-based \cite{Karras2019AnalyzingAI} encoder and decoder network for this image translation.
%
However, in contrast to Animatable Gaussians \cite{Li2023AnimatableGL}, we propose a more lightweight image translation pipeline, where we reduce the number of encoders from three to two and the number of decoders from six to two, and decrease the size of the StyleGAN \cite{Karras2019AnalyzingAI} decoder.
To efficiently use the 2D map space, we do not use projective textures from meshes~\cite{Wang2023StyleAvatarRP}, but a UV parametrization which reduces half of the decoders in comparison to StyleAvatar.
Moreover, we do not use triplets of the encoder-decoder, as we combine all properties of the Gaussians into one map $\pmb{G}_i$ following ASH~\cite{Pang2023ASHAG}. Therefore, we define our network as follows:
\begin{equation}
\begin{split}
\label{formula: appearance model}
\pmb{D}_i &: \Pi(\phi(\Gamma_{\text{PCA}}(\pmb{M}_i))), \\
\pmb{C} &: \Psi(\phi(\pmb{M})), \\
\end{split}
\end{equation}
where \mbox{$\pmb{D}_i \in \mathbb{R}^{11 \times H \times W}$} is a map containing the delta for positions $\Delta_{pos} \in \mathbb{R}^3$, rotation $\Delta_{rot} \in \mathbb{R}^4$, scale $\Delta_{scale} \in \mathbb{R}^3$, and opacity $\Delta_{op} \in \mathbb{R}$.
The colors \mbox{$\pmb{C} \in \mathbb{R}^{3 \times H \times W}$} of the Gaussians are predicted using the normal map $\phi(\pmb{M})$ of the canonical mesh $\pmb{M}$ ($\phi$ is the normal map extractor from a mesh).
Similar to Animatable Gaussians~\cite{Li2023AnimatableGL}, we use a PCA layer $\Gamma_{\text{PCA}}$ which serves as a low-pass regularization filter for the input. 
$\Gamma_{\text{PCA}}$ is built by using PCA on the meshes $\pmb{M}_i$ for the training frames and 16 principle components are used as the basis.
For the training, we use the projection of each incoming mesh $\pmb{M}_i$ on the PCA manifold 
\mbox{$\pmb{M}_{PCA_i}=\Gamma_{\text{PCA}}(\pmb{M}_i)$}. Optionally we use a feature texture that is concatenated with rasterized normals for the $\pmb{D}_i$ conditioning.

The final Gaussian map \mbox{$\pmb{G}_i \in \mathbb{R}^{11 \times H \times W}$} is obtained by applying the deltas to the canonical Gaussians $\pmb{G}$ \cite{Li2023AnimatableGL, Zielonka2025Drivable3D}. 
The deformed position of 3D means is computed as \mbox{$\pmb{T}_i(\pmb{M} + \pmb{D}_{i_{0:3}})$}. 
One major distinction compared to AG \cite{Li2023AnimatableGL} is the way how the transformation from the canonical space to the deformed space is handled. We employ deformation gradients, following Sumner et al.~\cite{Sumner2004DeformationTF}.
This approach allows for greater flexibility regarding input meshes, provided they maintain full correspondence.

Given a mesh $\pmb{M}_{PCA_i}$ for the frame $i$, we define the deformation gradients as $\mathbf{J}_j = \hat{\mathbf{E}}_j \mathbf{E}_j^{-1}$, where $\hat{\mathbf{E}}_j \in \mathbb{R}^{3\times3}$ and $\mathbf{E}_j \in \mathbb{R}^{3\times3}$ contain the Frenet frame (tangent, bi-tangent, normal) of the triangle $j$ defined in deformed and canonical spaces, respectively. Using these deformation gradients and the known correspondences between the Gaussian map and the meshes, we transform the Gaussians from the canonical space to the deformed space.

Note that our color map \pmb{C} is static and does not model view-dependent effects; this means that we force the network to recover globally consistent colors for each Gaussian similar to a texture in the classic 3DMM. Therefore, $\pmb{G}_i$ must model the wrinkles and self-shadows.
Finally, we use Gaussian splatting~\cite{Kerbl20233DGS} to render the regressed Gaussian maps.
We define the predicted color of pixel $(u,v)$ as:
\begin{equation}
\label{formula: splatting&volume rendering}
    \mathbf{\bar{C}}_{u,v} = \sum_{i\in \mathcal{N}}\mathbf{c_i} \alpha_i \prod_{j=1}^{i-1} (1-\alpha_j),
\end{equation}
where $\mathbf{c}_i$ is the Gaussian color predicted by $\Psi$, $\mathcal{N}$ is the number of texels and $\alpha_i$ is predicted opacity per Gaussian.

\subsection{Image-based Coefficents Regressor}

\begin{figure*}[ht!]
    \centering
    \setlength{\unitlength}{0.1\linewidth}
    \begin{picture}(11, 7.8)
    \put(0, 0.0){\includegraphics[width=1\linewidth]{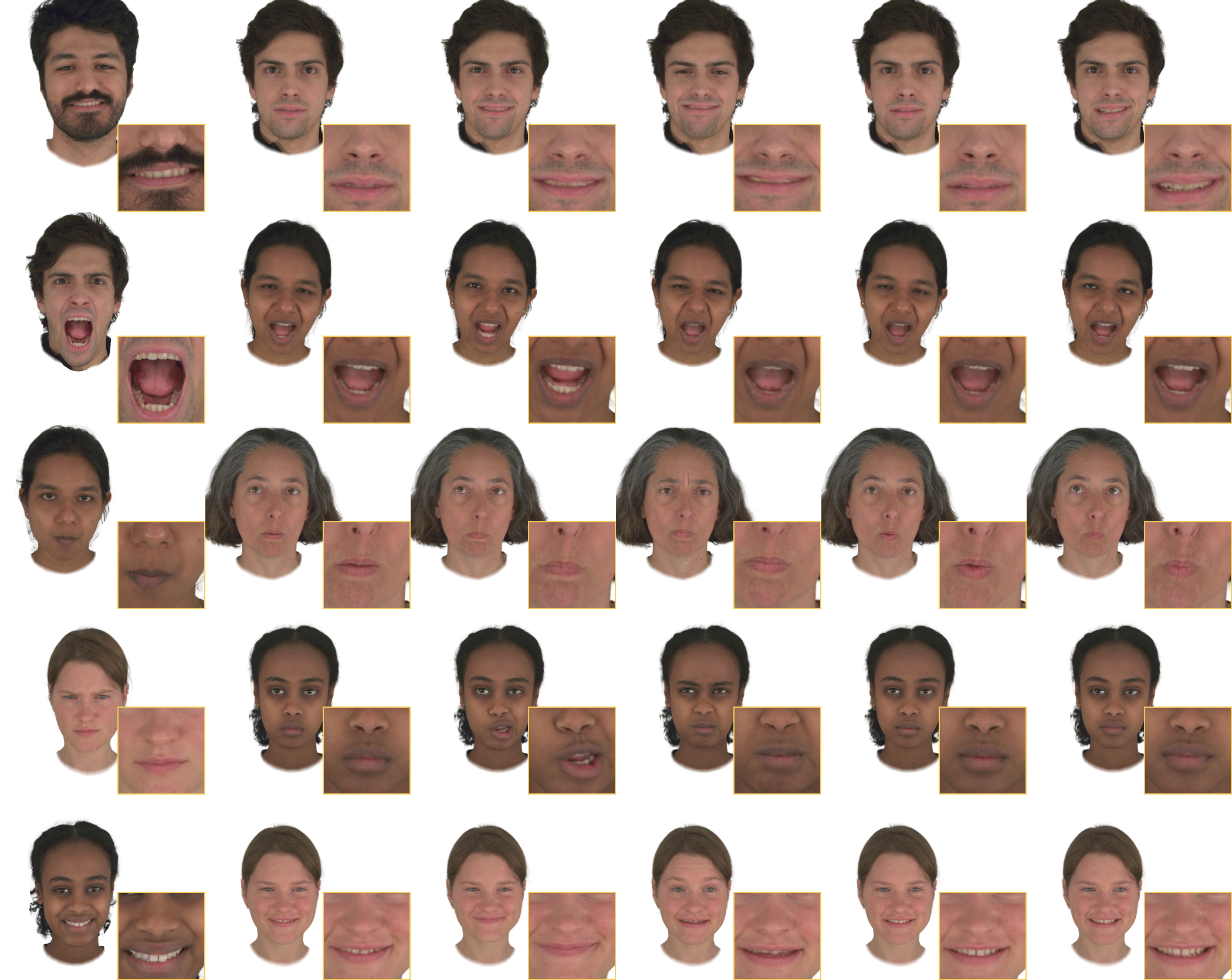}}
    \put(0.45, -0.35){Source}
    \put(2.1, -0.35){Relative}
    \put(4.0, -0.35){DECA \cite{Feng2020LearningAA}}
    \put(5.4, -0.35){Absolute}
    \put(7.1, -0.35){EMOCA \cite{Danek2022EMOCAED}}
    \put(8.8, -0.35){FLAME \cite{FLAME:SiggraphAsia2017}}
    \end{picture}
    \vspace{0.3cm}
    \caption{Our experiments show that using pre-trained regressor like EMOCA \cite{Danek2022EMOCAED} and DECA \cite{Feng2020LearningAA} work well for driving our GEM model. In this context, DECA and EMOCA refer to using either of the regressed feature vectors. FLAME represents the expression vectors regressed by EMOCA. 
\textit{Relative} and \textit{absolute} denote whether the EMOCA + DECA features are used directly or as relative changes from a neutral face.}
    \label{fig:regressor_ablation}
    \vspace{-0.4cm}
\end{figure*}

Table \ref{tab:regressor_ablation} shows an ablation study in the context of input to our MLP regressor which predicts GEM coefficients. Figure \ref{fig:regressor_ablation} provides an additional qualitative comparison. Using a pre-trained decoder such as EMOCA demonstrates strong potential for cross-reenactment by leveraging robust priors. Future work will explore further methods for image-based control of GEM, with a potential approach being the incorporation of additional modalities, such as sound, into the EMOCA-based regressor.

\begin{table}[t]
  \centering
  \resizebox{0.8\columnwidth}{!}{
    \begin{tabular}{l|rrrr}
    \toprule
    \textbf{Ablation} & L1 $\downarrow$  & LIPIS $\downarrow$ & PSNR  $\uparrow$ & SSIM $\uparrow$ \\
    \midrule
    Ours      & 0.0181 & 0.1171 & 24.2997 & \textbf{0.9134} \\
    Absolute  & 0.0181 & 0.1172 & \textbf{24.3065} & 0.9130 \\
    FLAME      & 0.0181 & 0.1169 & 24.2910 & 0.9130 \\
    EMOCA \cite{Danek2022EMOCAED}  & \textbf{0.0180} & \textbf{0.1165} & 24.2715 & 0.9132 \\
    DECA \cite{Feng2020LearningAA}  & 0.0184 & 0.1189 & 24.2181 & 0.9127 \\
    \bottomrule
    \end{tabular}
  }
   \caption{We performed an ablation study of our regressor using self-reenactment tasks. In this study, we tested configurations that used only EMOCA or DECA features, as well as a version where the EMOCA regressed FLAME expressions. Lastly, we show the effect of using absolute features instead of the relative ones used in Ours (features relative to the neutral face).}
  \label{tab:regressor_ablation}
  \vspace{-0.1cm}
\end{table}

\subsection{CNN Training Details}
The training objective of the CNN-based appearance model is defined as $\mathcal{L} = \mathcal{L}_{Color} + \mathcal{L}_{Reg}$.
$\mathcal{L}_{Color}$ is a weighted sum of three different photo-metric losses between the rendered image $\mathbf{\bar{C}}$ and the ground truth $\mathbf{C}$:
\begin{equation}
\begin{split}
\label{formula: color loss}
\mathcal{L}_{Color} &= 
        (1 - \omega) \mathcal{L}_1 + \omega\mathcal{L}_{\text{D-SSIM}} + \zeta\mathcal{L}_{\text{VGG}}, \\
\mathcal{L}_{Reg} &= \lambda \sum_{j=1}^{N} \norm{\Delta_{pos_j}}^2 + \gamma \sum_{j=1}^{N} \norm{s_{scale_j}}^2, \\
\end{split}
\end{equation}
where $\omega=0.2$, $\zeta=0.0075$ (after 150k iterations steps and zero otherwise), $\mathcal{L}_{\text{D-SSIM}}$ is a structural dissimilarity loss, and $\mathcal{L}_{\text{VGG}}$ is the perceptual VGG loss.
$\mathcal{L}_{Reg}$ regularizes position offsets $\Delta_{pos_j}$ and scales $s_{scale_j}$ to stay small w.r.t. the input mesh.
We train our model for $10^6$ steps using Adam \cite{Kingma2014AdamAM} with a learning rate $5\mathrm{e}{-4}$ and a batch size of one which takes around 10h on a Nvidia RTX4090.

\begin{figure*}[th!]
    \centering
    \centering
    \setlength{\unitlength}{0.1\textwidth}
    \begin{picture}(10, 2.5)
    \put(-0.2, 0){\includegraphics[width=1.0\textwidth]{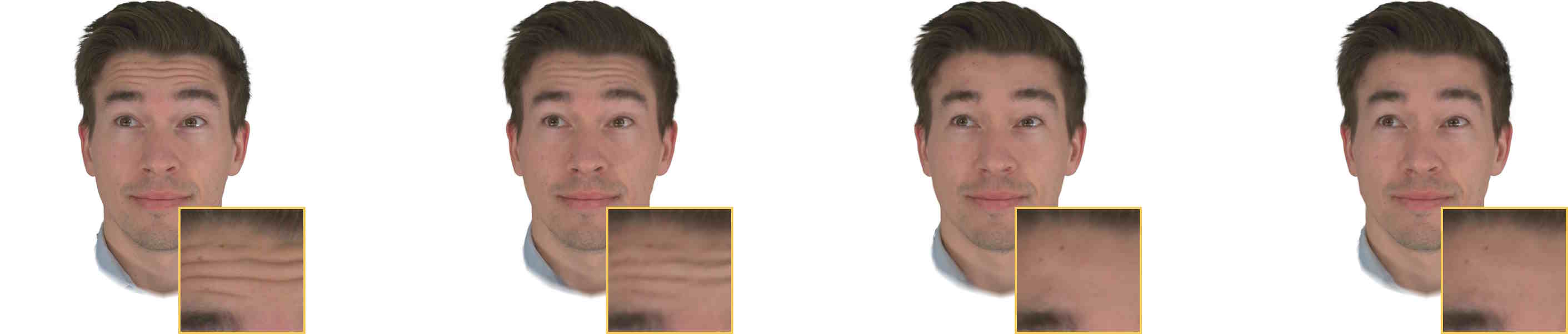}}
    \put(0.3, -0.4){Ground Truth}
    \put(2.6, -0.4){Ours \#30 $128^2$}
    \put(5.7, -0.4){GA \cite{Qian2024gaussianavatars} $128^2$}
    \put(8.4, -0.4){GA \cite{Qian2024gaussianavatars} $256^2$}
    \end{picture}
    \vspace{0.1cm}
    \caption{The quality comparison to Gaussian Avatars \cite{Qian2024gaussianavatars} shows better performance, even though our model is similar in size to its Gaussian cloud, and we do 
    \textbf{not} need an additional FLAME model, which weighs 90 MB.}
    \label{fig:ga_avatar_size_comparison_supp}
    \vspace{-0.8cm}
\end{figure*}

\textbf{Our method and all the baselines were trained using the same multiview input data sourced from the dataset provided by Qian} \cite{Qian2024gaussianavatars}, which includes multiview images from the NeRSamble dataset \cite{Kirschstein2023NeRSembleMR} as well as tracked meshes.

\section{Compression Ablation Study}

In the domain of 3D Morphable Models, principle component analysis (PCA) emerges as a cornerstone approach, instrumental in crafting the foundational framework for capturing face expressions and shapes with remarkable fidelity \cite{Blanz1999AMM, FLAME:SiggraphAsia2017}.
This methodology has been adopted with notable success, not only in modeling facial features but also in extrapolating the nuances of human bodies \cite{SMPL-X:2019, SUPR:2022, Loper2015SMPL}, and even in depicting intricate hand modeling \cite{MANO:SIGGRAPHASIA:2017}.

\begin{figure*}[ht!]
    \centering
    \vspace{-0.1cm}
    \centering
    \setlength{\unitlength}{0.1\textwidth}
    \begin{picture}(10, 8)
    \put(-0.1, 0){\includegraphics[width=1.0\textwidth]{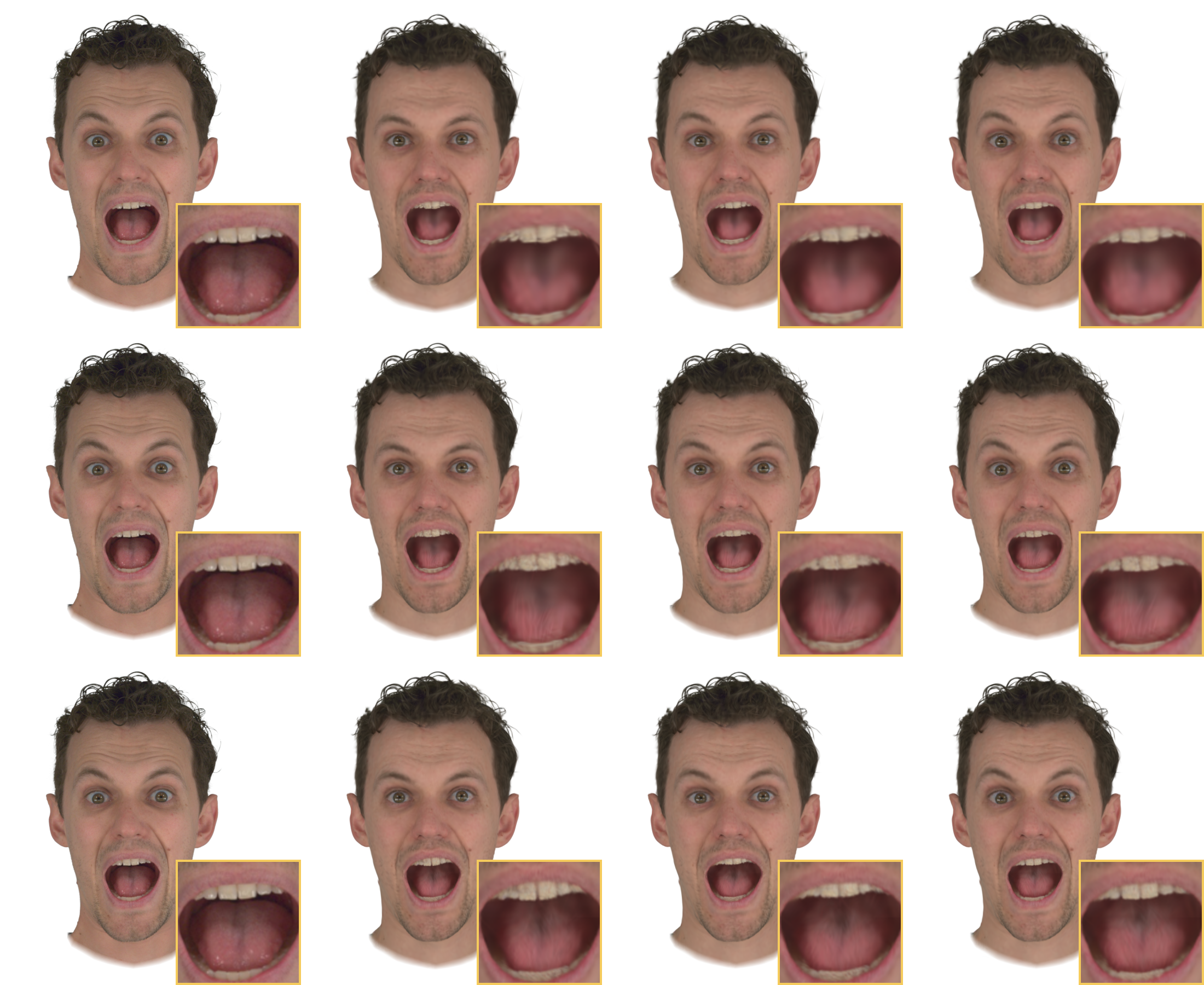}}
    \put(-0.1, 3.1){\rotatebox{90}{Ground Truth}}
    \put(9.9, 6.2){\rotatebox{90}{$128^2$}}
    \put(9.9, 3.5){\rotatebox{90}{$256^2$}}
    \put(9.9, 0.8){\rotatebox{90}{$512^2$}}
    \put(3.15, -0.2){\#10}
    \put(5.65, -0.2){\#30}
    \put(8.15, -0.2){\#50}
    \end{picture}
    \vspace{0.1cm}
    \caption{Qualitative compression quality depending on the number of basis (10, 30, 50) and resolution of the Gaussian map ($128^2$, $256^2$, $512^2$).}
    \label{fig:complete_compression}
    \vspace{-0.6cm}
\end{figure*}

\begin{figure*}[ht!]
    \centering
    \setlength{\unitlength}{0.1\textwidth}
    \begin{picture}(10, 10)
    \put(1.0, 0){\includegraphics[width=0.9\textwidth]{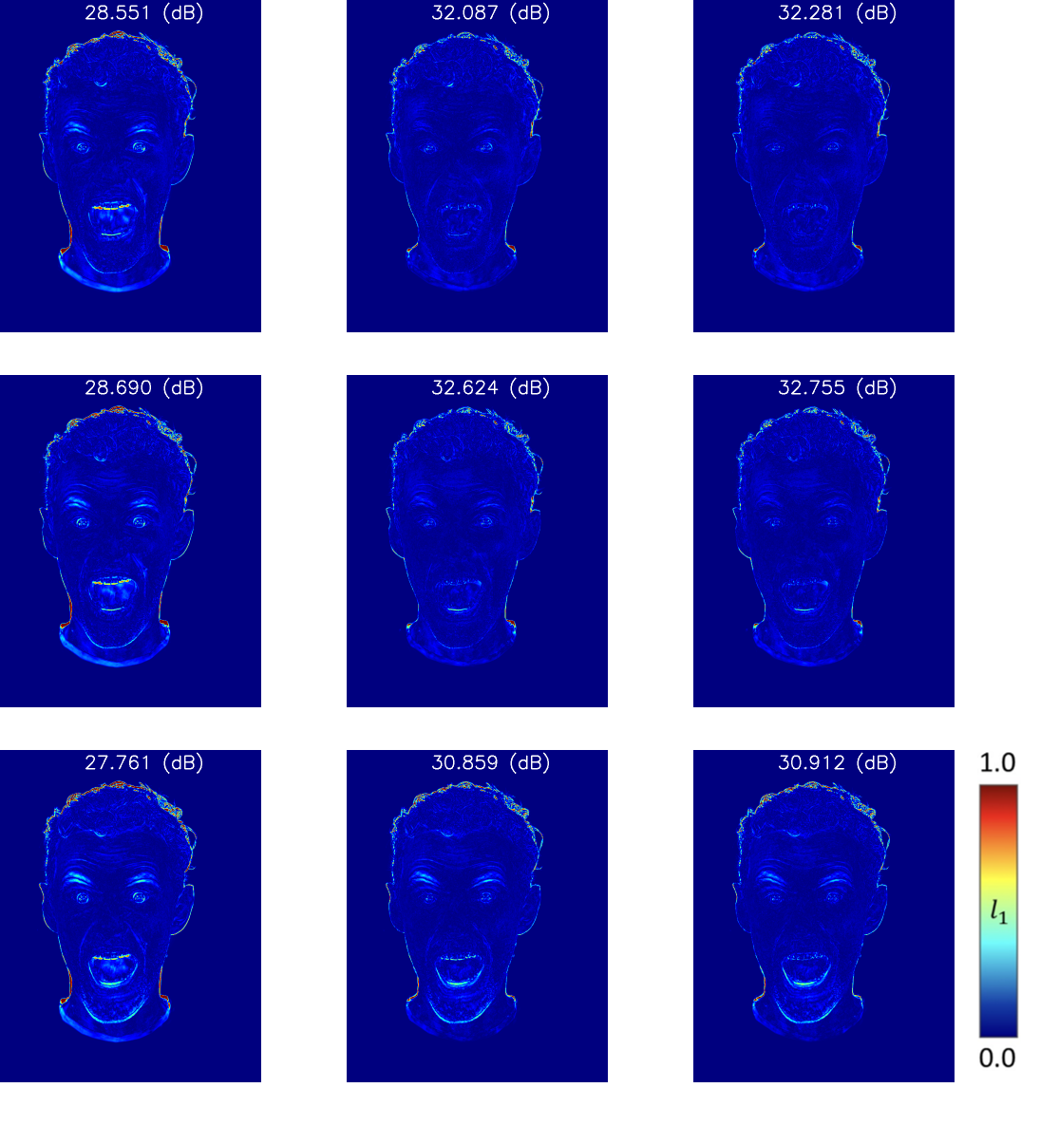}}
    \put(0.3, 8.2){\rotatebox{90}{$128^2$}}
    \put(0.3, 5.0){\rotatebox{90}{$256^2$}}
    \put(0.3, 1.5){\rotatebox{90}{$512^2$}}
    \put(1.8, -0.2){\#10}
    \put(4.8, -0.2){\#30}
    \put(7.8, -0.2){\#50}
    \end{picture}
    \vspace{-0.1cm}
    \caption{Compression error in PSNR (db) depending on the number of basis (10, 30, 50) and resolution of the Gaussian maps ($128^2$, $256^2$, $512^2$).}
    \label{fig:complete_heatmap}
\end{figure*}

\begin{figure*}[ht!]
    \centering
    \vspace{-0.1cm}
    \centering
    \setlength{\unitlength}{0.1\textwidth}
    \begin{picture}(10, 8)
    \put(-0.1, 0){\includegraphics[width=1.0\textwidth]{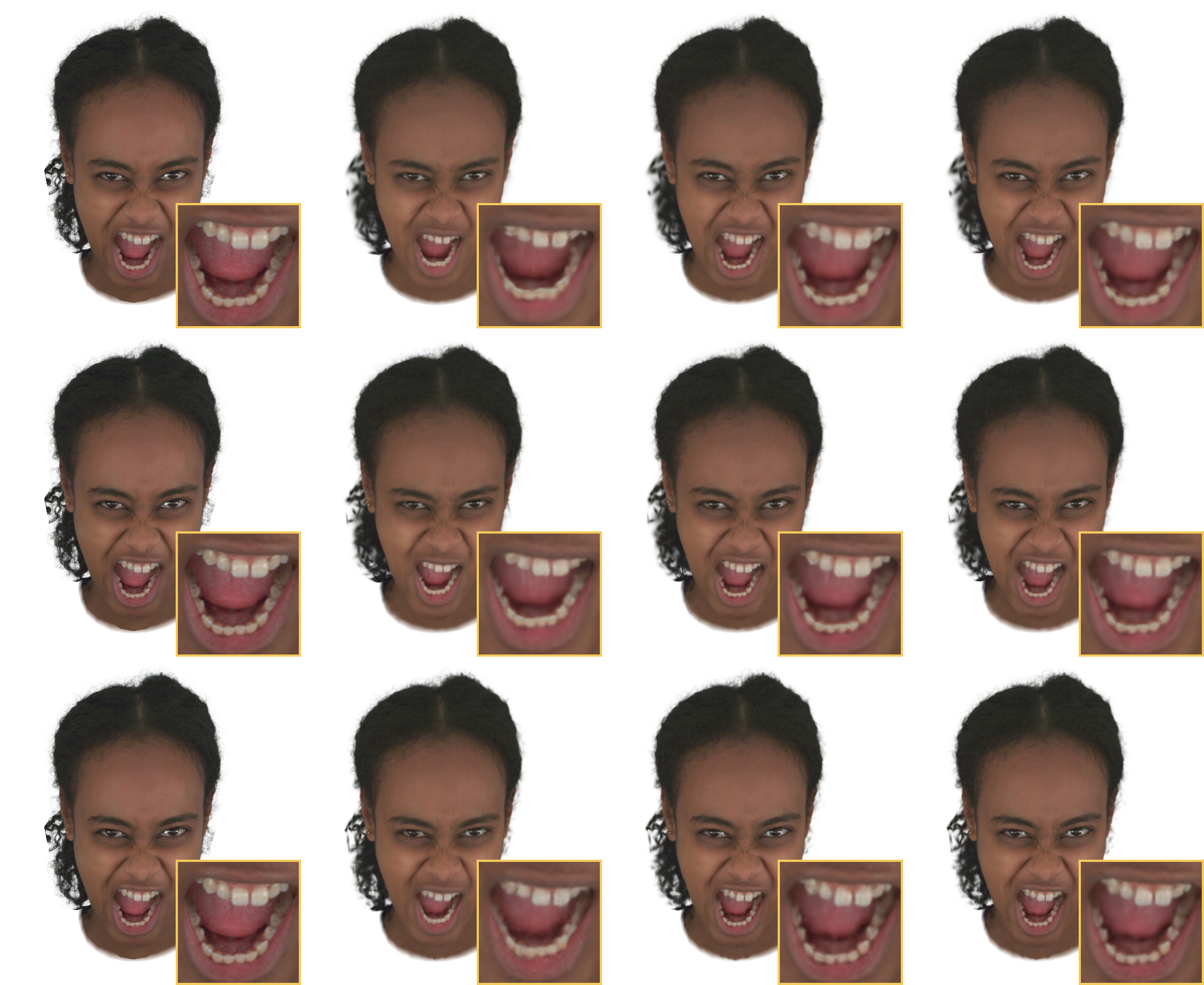}}
    \put(-0.1, 3.1){\rotatebox{90}{Ground Truth}}
    \put(9.9, 6.2){\rotatebox{90}{$128^2$}}
    \put(9.9, 3.5){\rotatebox{90}{$256^2$}}
    \put(9.9, 0.8){\rotatebox{90}{$512^2$}}
    \put(3.15, -0.2){\#10}
    \put(5.65, -0.2){\#30}
    \put(8.15, -0.2){\#50}
    \end{picture}
    \vspace{0.1cm}
    \caption{Qualitative compression quality depending on the number of basis (10, 30, 50) and resolution of the Gaussian map ($128^2$, $256^2$, $512^2$).}
    \label{fig:complete_compression_1}
    \vspace{-0.6cm}
\end{figure*}

\begin{figure*}[ht!]
    \centering
    \setlength{\unitlength}{0.1\textwidth}
    \begin{picture}(10, 10)
    \put(1.0, 0){\includegraphics[width=0.9\textwidth]{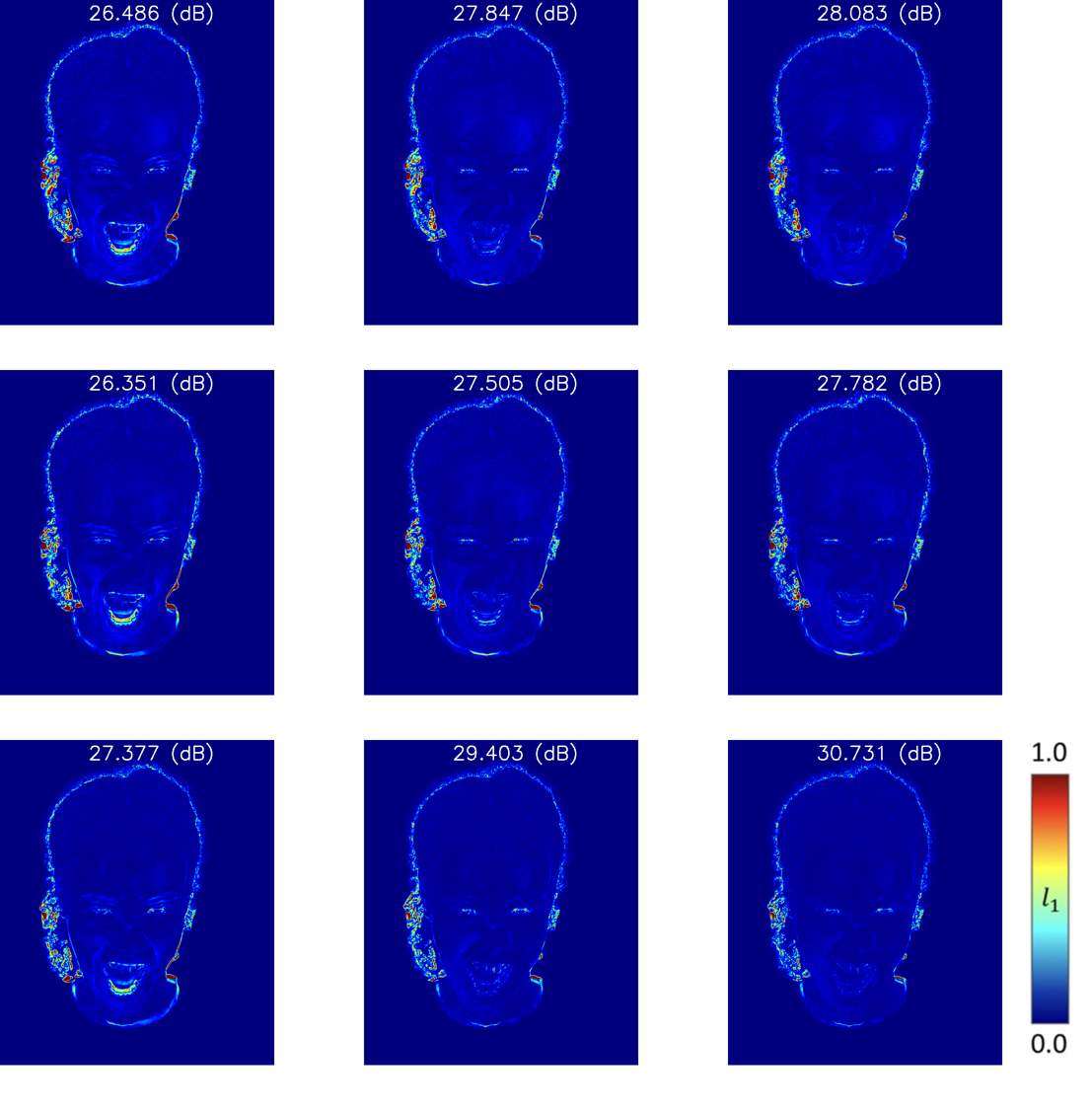}}
    \put(0.3, 7.7){\rotatebox{90}{$128^2$}}
    \put(0.3, 4.7){\rotatebox{90}{$256^2$}}
    \put(0.3, 1.5){\rotatebox{90}{$512^2$}}
    \put(1.8, -0.2){\#10}
    \put(4.8, -0.2){\#30}
    \put(7.8, -0.2){\#50}
    \end{picture}
    \vspace{-0.1cm}
    \caption{Compression error in PSNR (db) depending on the number of basis (10, 30, 50) and resolution of the Gaussian maps ($128^2$, $256^2$, $512^2$).}
    \label{fig:complete_heatmap_1}
\end{figure*}

\begin{figure*}[ht!]
    \centering
    \vspace{-0.1cm}
    \centering
    \setlength{\unitlength}{0.1\textwidth}
    \begin{picture}(10, 8)
    \put(-0.1, 0){\includegraphics[width=1.0\textwidth]{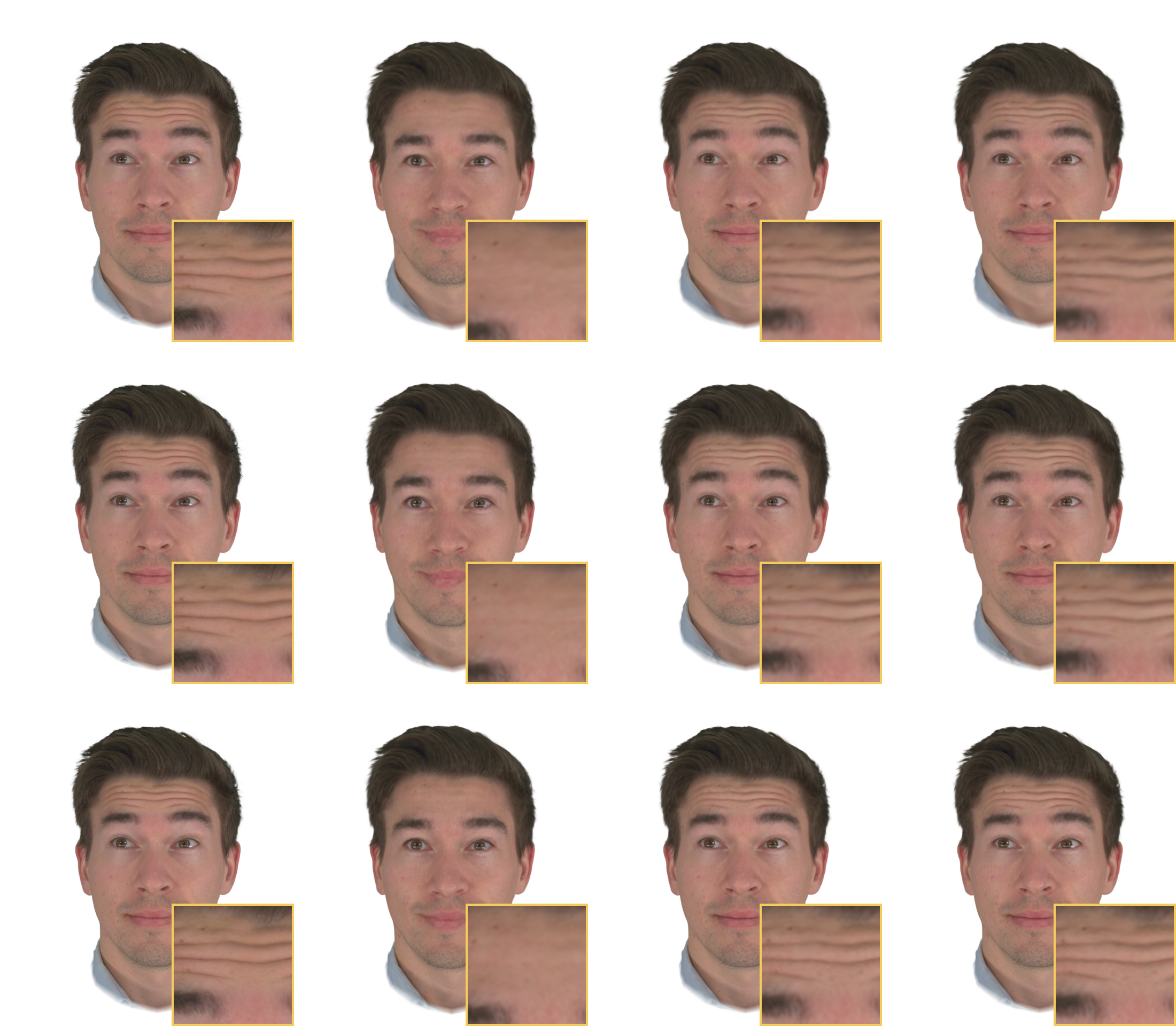}}
    \put(-0.1, 3.1){\rotatebox{90}{Ground Truth}}
    \put(9.9, 6.2){\rotatebox{90}{$128^2$}}
    \put(9.9, 3.5){\rotatebox{90}{$256^2$}}
    \put(9.9, 0.8){\rotatebox{90}{$512^2$}}
    \put(3.15, -0.2){\#10}
    \put(5.65, -0.2){\#30}
    \put(8.15, -0.2){\#50}
    \end{picture}
    \vspace{0.1cm}
    \caption{Qualitative compression quality depending on the number of basis (10, 30, 50) and resolution of the Gaussian map ($128^2$, $256^2$, $512^2$).}
    \label{fig:complete_compression_0}
    \vspace{-0.6cm}
\end{figure*}

\begin{figure*}[ht!]
    \centering
    \setlength{\unitlength}{0.1\textwidth}
    \begin{picture}(10, 10)
    \put(1.0, 0){\includegraphics[width=0.9\textwidth]{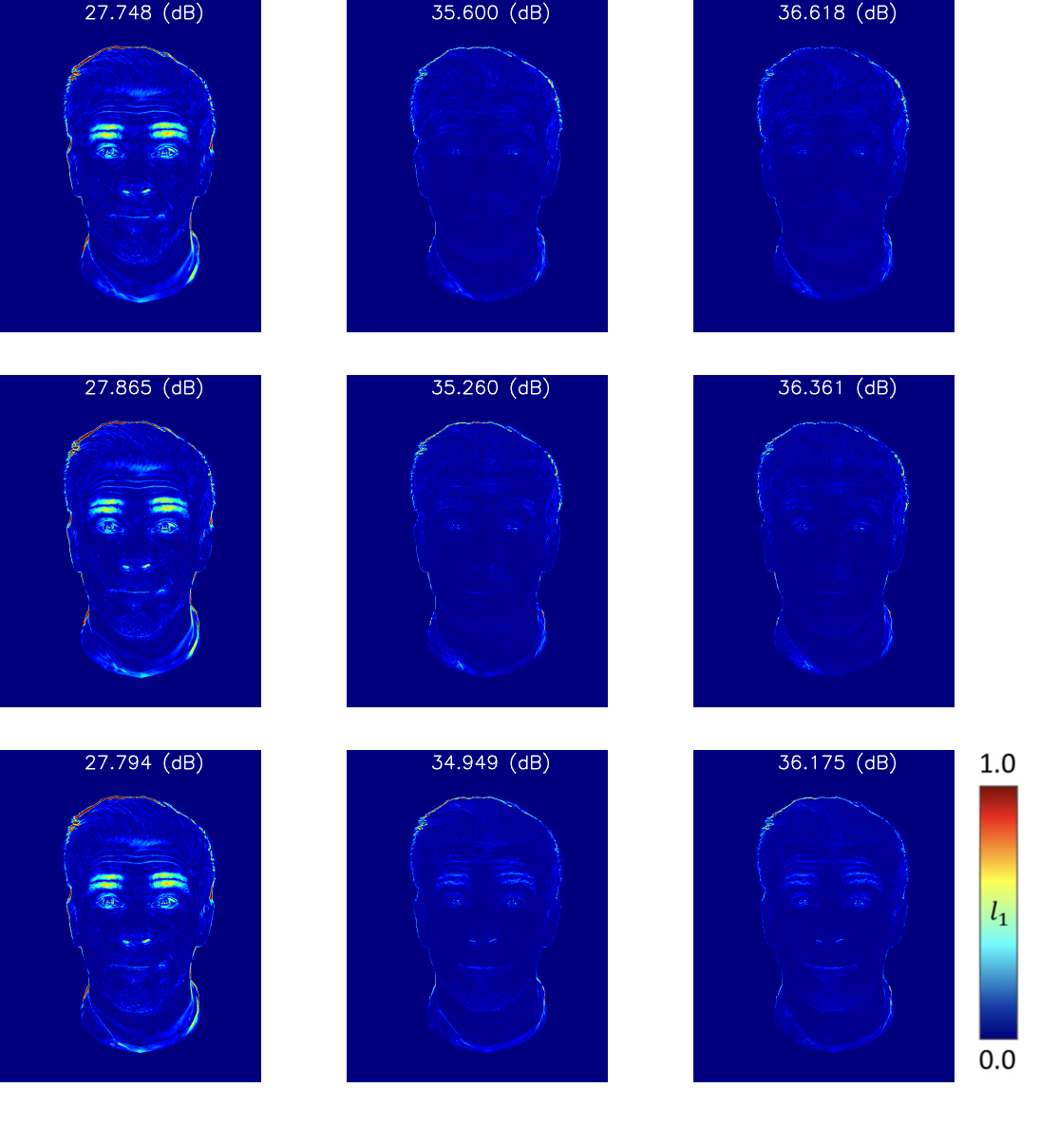}}
    \put(0.3, 8.2){\rotatebox{90}{$128^2$}}
    \put(0.3, 5.0){\rotatebox{90}{$256^2$}}
    \put(0.3, 1.5){\rotatebox{90}{$512^2$}}
    \put(1.8, -0.2){\#10}
    \put(4.8, -0.2){\#30}
    \put(7.8, -0.2){\#50}
    \end{picture}
    \vspace{-0.1cm}
    \caption{Compression error in PSNR (dB) depending on the number of bases (10, 30, 50) and resolution of the Gaussian maps ($128^2$, $256^2$, $512^2$).}
    \label{fig:complete_heatmap_0}
\end{figure*}

Expanding upon this foundation, \shortname proposes a novel technique involving an ensemble of eigenbases of 3D Gaussian attributes for achieving a photorealistic human head appearance.
This representation exhibits significant adaptability concerning both quality and size, leveraging a fundamental trait of linear basis that proves beneficial when applied to diverse devices with varying capabilities in digital human applications.
Figures \ref{fig:complete_compression} and \ref{fig:complete_heatmap} illustrate the qualitative and quantitative results of \shortname.
Notably, even under substantial compression (utilizing only ten principal components), our approach consistently yields high-quality outcomes. More examples can be found in Figures \ref{fig:complete_compression_1}, \ref{fig:complete_heatmap_1}, \ref{fig:complete_compression_0} and \ref{fig:complete_heatmap_0}.

\section{Human Head Avatar Compression}

Human avatar compression is an important topic, but it is still in its early stages and not well-explored.
For neural-based representations, there are methods to compress networks, such as pruning \cite{han2015learningweightsconnectionsefficient}, quantization \cite{jacob2017quantizationtrainingneuralnetworks}, or knowledge distillation \cite{hinton2015distillingknowledgeneuralnetwork}, as well as small and compact MobileNets \cite{howard2017mobilenetsefficientconvolutionalneural}.
Interestingly, in the latter context, GEM can be considered as a single-layer MLP without any activation function.
Unfortunately, these methods still require an expensive forward pass and may not be well-suited for all commodity devices.

\begin{table}[h]
  \centering
  \resizebox{0.6\linewidth}{!}{
    \begin{tabular}{l|r|r|r}
    \#Comp & $128^2$ & $256^2$ & $512^2$ \\
    \midrule
    10 & 7  & 28 & 113  \\
    30 & 20 & 83 & 333  \\
    50 & 34 & 138 & 553 \\
    \bottomrule
    Ours Net     & 82 & 109 & 178 \\
    GA StyleUnet & 487 & 529 & 636 \\
    \bottomrule
    \end{tabular}
  }
  \caption{Memory consumption (in MB with float32) of GEM depends on the texture resolution and number of components. Our model shows much better granularity compared to the fixed size of neural networks and can be adjusted on the fly depending on the hardware.}
  \label{tab:pca_size_mb}
\end{table}

\begin{table}[h]
  \centering
  \vspace{-0.75cm}
  \resizebox{0.6\linewidth}{!}{
    \begin{tabular}{l|r|r|r}
    \#Comp & $128^2$ & $256^2$ & $512^2$ \\
    \midrule
    10 & 31.47 & 31.90 & 31.80  \\
    30 & 33.46 & 34.30 & 34.26  \\
    50 & 33.79 & 34.75 & 34.73 \\   
    \bottomrule
    \end{tabular}
  }
  \caption{PSNR color error in dB for one actor with a different number of principle components and Gaussian map resolutions. Despite heavy compression (10 principal components), the avatar is still of high quality. More details are in the supplementary material.
  }
  \vspace{-0.2cm}
  \label{tab:pca_compression}
\end{table}

\begin{figure*}[hb!]
    \centering
    \setlength{\unitlength}{0.1\linewidth}
    \begin{picture}(10, 4.2)
    \put(1.6, 0.0){\includegraphics[width=0.7\linewidth]{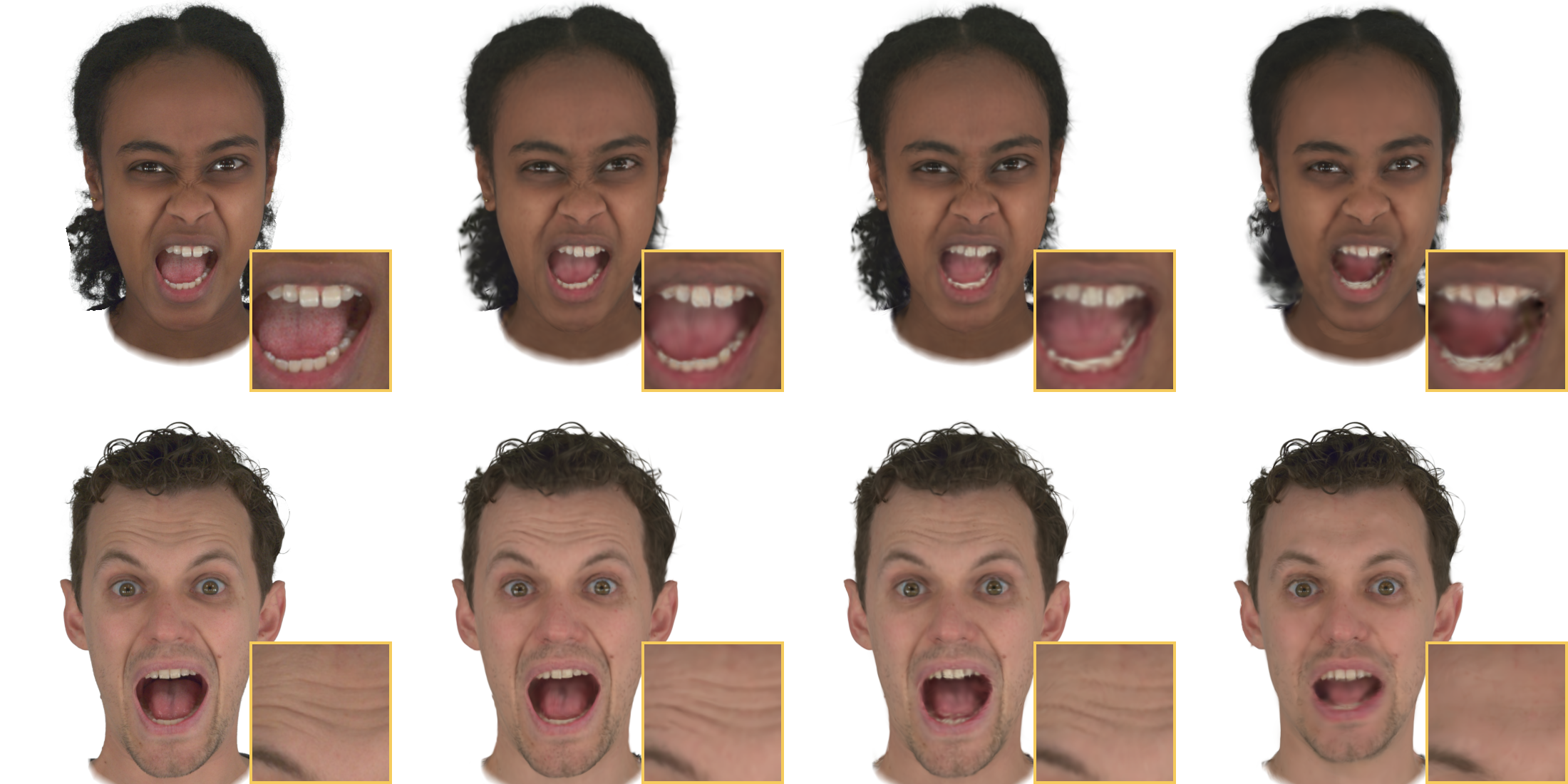}}
    \put(2.2, -0.25){Ground Truth}
    \put(4.3, -0.25){GEM}
    \put(5.6, -0.25){GEM on AG}
    \put(7.5, -0.25){GEM on GA}
    \end{picture}
    \vspace{0.1cm}
    \caption{GEM applied on different avatar methods (AG and GA) and optimized using analysis-by-synthesis. Our method is universal and can be successfully used on point clouds and textures to distill a lightweight avatar.}
    \label{fig:generalizability}
    \vspace{-0.1cm}
\end{figure*}

The recently introduced Gaussian Avatars by Qian \cite{Qian2024gaussianavatars} also represents a form of avatar compression, though not in the primitives' space but rather in the geometry space, with Gaussians attached and deformed by triangles from a linear face model.
However, this form of appearance representation is insufficient for capturing details such as wrinkles, as it rigidly adheres to FLAME rigging in the geometry space.
Therefore, we advocate for different compression techniques like GEM, which can leverage more powerful representations and distill them into expressive, high-quality linear models.
We hope that this project will open doors to different methods for efficiently storing and representing avatars.

\begin{figure*}[b!]
    \centering
    \setlength{\unitlength}{0.1\linewidth}
    \begin{picture}(10, 3)
    \put(0, 0.0){\includegraphics[width=1\linewidth]{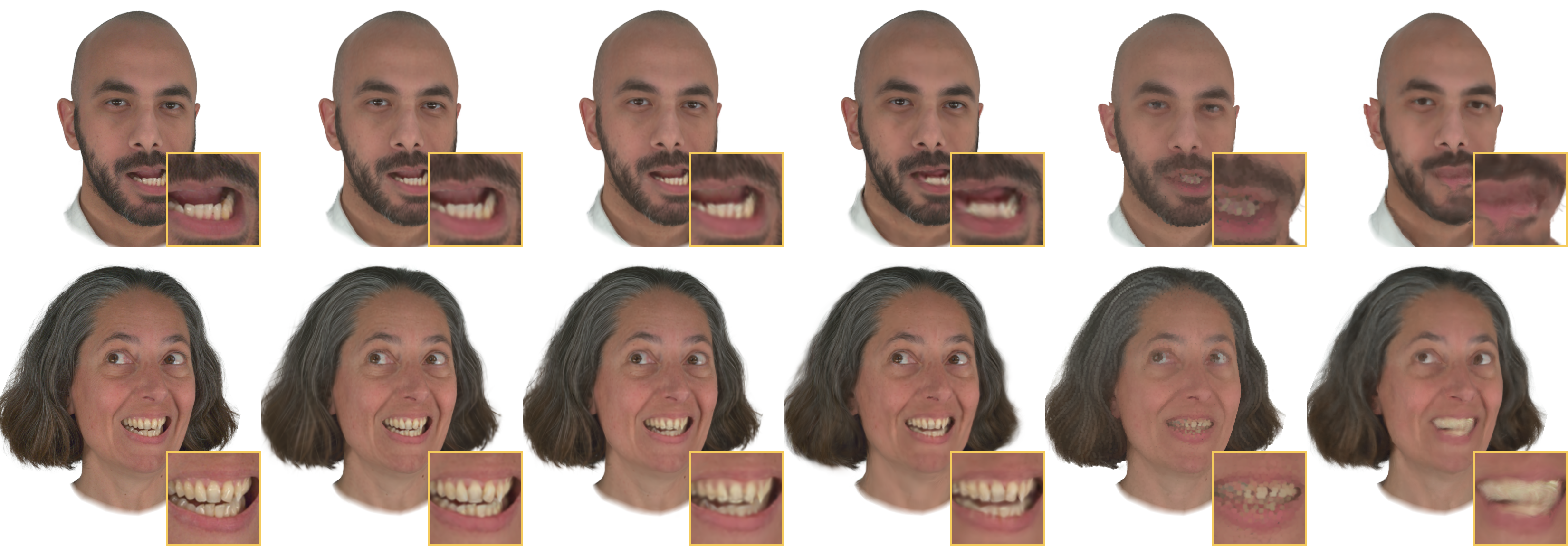}}
    \put(0.15, -0.35){Ground Truth}
    \put(2.1, -0.35){Ours GEM}
    \put(4.0, -0.35){AG \cite{Li2023AnimatableGL}}
    \put(5.4, -0.35){GA \cite{Qian2024gaussianavatars}}
    \put(6.6, -0.35){PointAvatar \cite{Zheng2022PointAvatarDP}}
    \put(8.8, -0.35){AvatarMAV\cite{xu2023avatarmav}}
    \end{picture}
    \vspace{0.2cm}
    \caption{Additional baselines PointAvatar (PSNR: 25.8, SSIM: 0.893 LPIPS: 0.097) and AvatarMAV (PSNR: 29.5, SSIM: 0.913, LPIPS: 0.152) evaluated on the novel-view sequences.}
    \label{fig:pointavatar}
    \vspace{-0.4cm}
\end{figure*}

\section{Additional Dataset Evaluation}
\begin{figure*}[ht!]
    \centering
    \centering
    \setlength{\unitlength}{0.1\textwidth}
    \begin{picture}(10, 9)
    \put(0, 0){\includegraphics[width=1.0\textwidth]{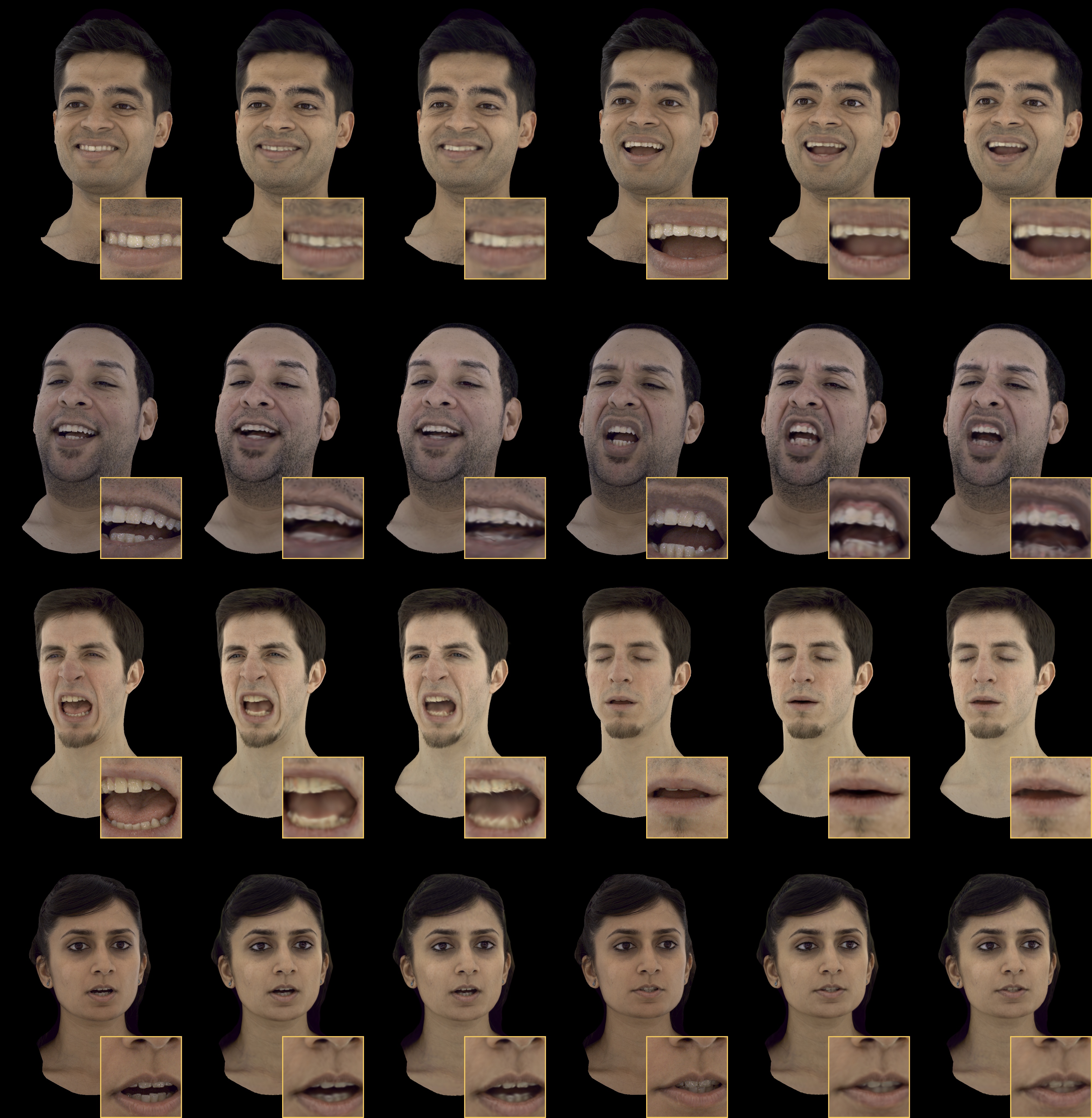}}
    \put(0.2, -0.4){Ground Truth}
    \put(2.2, -0.4){Ours Net}
    \put(3.7, -0.4){Ours GEM}
    \put(5.3, -0.4){Ground Truth}
    \put(7.3, -0.4){Ours Net}
    \put(8.7, -0.4){Ours GEM}
    \end{picture}
    \vspace{0.1cm}
    \caption{The Multiface dataset, introduced by Wu et al. \cite{wuu2022multiface}, comprises actors performing scripted expressions in short segments. A notable challenge arises due to the occurrence of several expressions, like "show all teeth," appearing only once in the dataset. This poses a difficulty during testing, particularly when the network is required to extrapolate. Here we showcase the outcomes of the test sequences to illustrate the effectiveness of our CNN-based network in capturing diverse and challenging facial poses, demonstrating its robustness despite the inherent complexity of the dataset.
    }
    \label{fig:multiface}
    \vspace{-0.6cm}
\end{figure*}

In Figure \ref{fig:multiface}, we provide further evaluation of our approach using the Multiface dataset \cite{wuu2022multiface}. This dataset encompasses short sequences of facial expressions, ranging from "relaxed mouth open" to "show all teeth" or "jaw open huge smile." The expressions vary widely in length and complexity, presenting a considerable challenge for analysis. It's important to note that this dataset does not provide a parametric 3DMM; instead, it offers meshes in full correspondence. However, as mentioned in the main text, our method remains adaptable in this context. By leveraging the deformation gradient \cite{Sumner2004DeformationTF} to transform points from canonical space into deformed space, and assuming consistent UV parametrization of input meshes, we can successfully navigate between these spaces. As depicted in Figure \ref{fig:multiface}, our network demonstrates the ability to extrapolate to novel expressions, even amidst highly challenging facial poses.

\section{Broader Impact}
Our project focuses on reconstructing a highly detailed human face avatar from multiview videos, enabling the extrapolation of expressions not originally captured. While our technology serves primarily constructive purposes, such as enriching telepresence or mixed reality applications, we acknowledge the potential risks of misuse. Therefore, we advocate for advancements in digital media forensics \cite{Rssler2019FaceForensicsLT, Rssler2018FaceForensicsAL} to aid in detecting synthetic media. Additionally, we emphasize the importance of conducting research in this area with transparency and openness, including the thorough disclosure of algorithmic methodologies, data origins, and models intended for research purposes.

\section{Future Applications \& Discussion}
\begin{figure*}[hb!]
    \centering
    \setlength{\unitlength}{0.1\textwidth}
    \begin{picture}(10, 5)
    \put(0.5, 0){\includegraphics[width=0.9\textwidth]{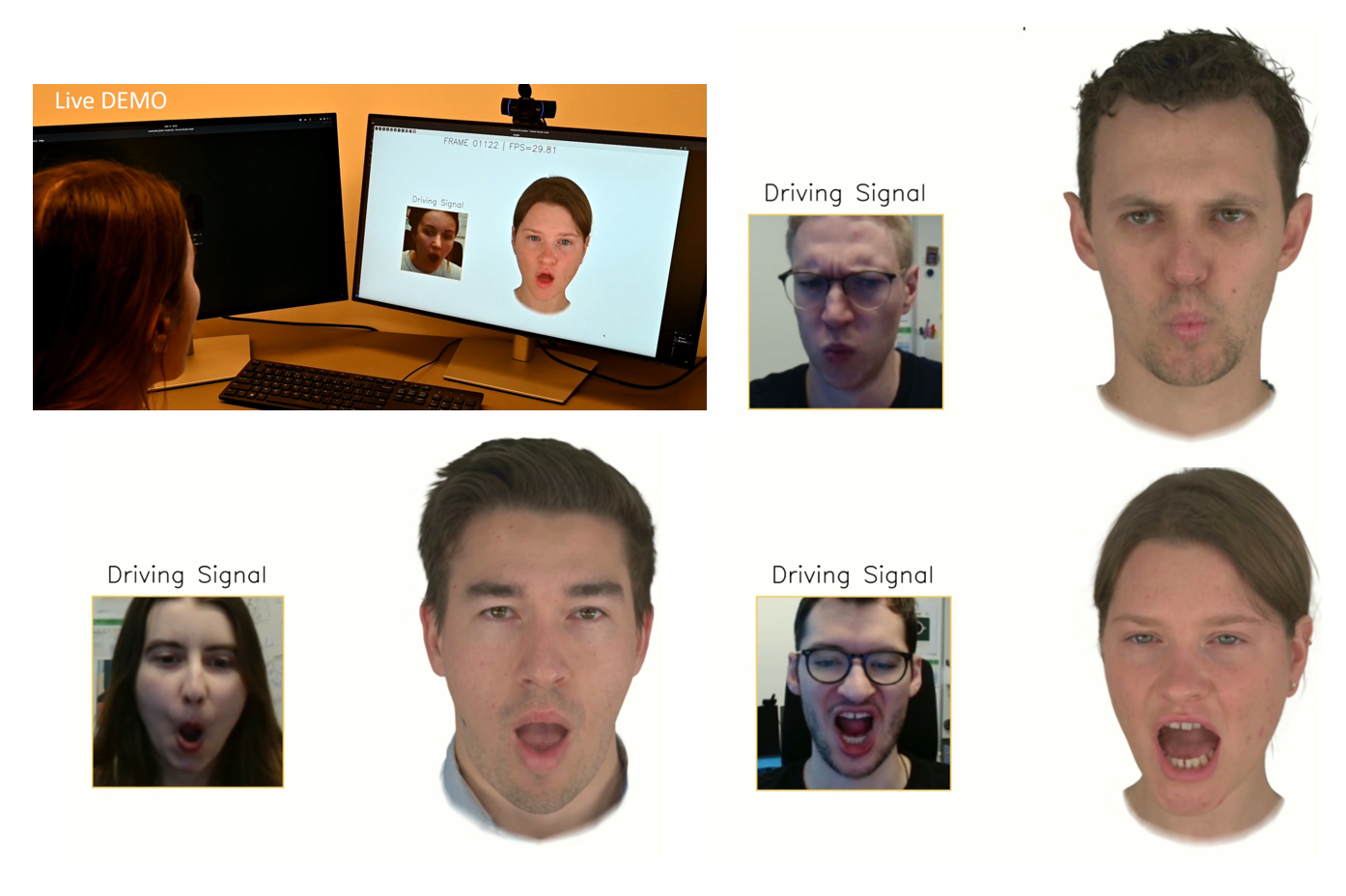}}
    \end{picture}
    \vspace{-1.0cm}
    \caption{GEM can be effectively controlled in real-time by an image-space regressor which produces coefficients projected on the linear basis of a personalized GEM avatar.}
    \label{fig:live_demo}
\end{figure*}

An interesting application venue for \shortname would be a combination of audio-driven methods with the appearance offered by our methods. Ng et at. \cite{Ng2024FromAT} presented photorealistic audio-driven full-body avatars. Despite impressive results, the face region still does not fully convey expressions and lacks realism. One way of improving it would be incorporating recent progress in audio-driven geometry \cite{Thambiraja2022ImitatorPS, Thambiraja20233DiFACEDS, Aneja2023FaceTalkAM, VOCA2019} with a dedicated appearance model offered by \shortname and our image-space regressor Figure \ref{fig:live_demo}.

Moreover, our neural network based appearance model uses meshes to obtain normal maps as input to the Gaussian map regressor (similar to the baselines). However, meshes are limited by resolution and expressiveness, one way of improving on that would be to use NPHM by Giebenhain et al. \cite{giebenhain2023nphm} and the follow-up work \cite{tang2024dphms, giebenhain2024mononphm, kirschstein2023diffusionavatars} to further increase the expressiveness of the model by explicitly capturing regions like hair or teeth.

\begin{figure*}[th!]
    \centering
    \vspace{-0.6cm}
    \setlength{\unitlength}{0.1\textwidth}
    \begin{picture}(10, 6)
    \put(0, 0){\includegraphics[width=\textwidth]{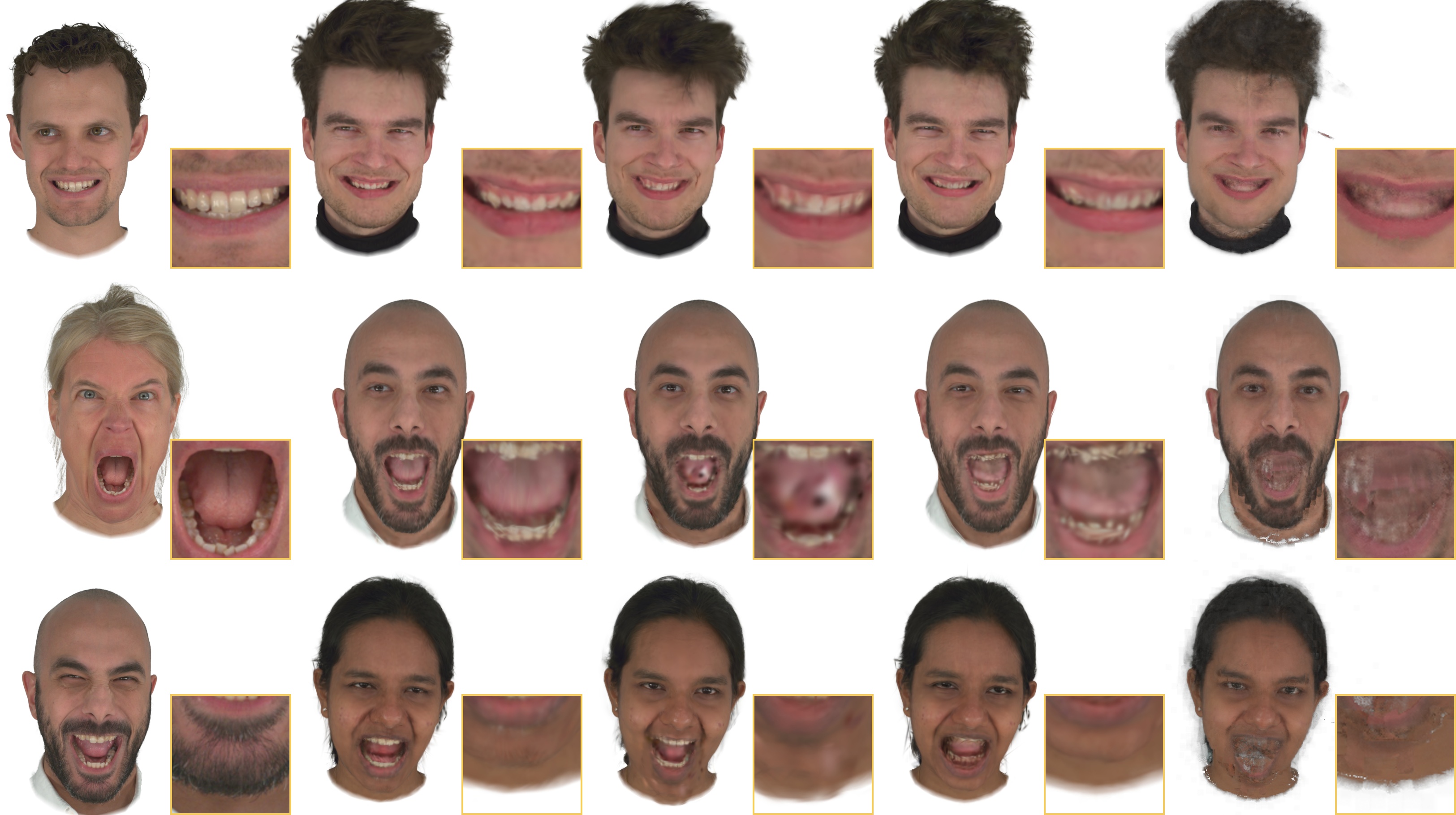}}
    \put(0.1, -0.4){Ground Truth}
    \put(2.5, -0.4){Ours Net}
    \put(4.5, -0.4){GA.~\cite{Qian2024gaussianavatars}}
    \put(6.6, -0.4){AG.~\cite{Li2023AnimatableGL}}
    \put(8.4, -0.4){INSTA~\cite{Zielonka2022InstantVH}}
    \end{picture}
    \vspace{0.25cm}
    \caption{\textbf{Facial cross-person reenactment.} The person's expressions on the left are transferred to the respective avatars on the right. In this experiment, we are using relative expressions based on ground truth meshes from the dataset (FLAME-based meshes reconstructed from multi-view data). Note that this experiment does not apply to our \shortname, since it is mesh-free.}
    \vspace{-0.5cm}
    \label{fig:transfer}
\end{figure*}

\end{document}